\documentclass{article}

 \usepackage[preprint]{neurips_2026}

\usepackage{marvosym}
\usepackage{float}

\usepackage[utf8]{inputenc} 
\usepackage[T1]{fontenc}    
\usepackage{hyperref}       
\usepackage{url}            
\usepackage{booktabs}       
\usepackage{amsfonts}       
\usepackage{nicefrac}       
\usepackage{microtype}      
\usepackage{xcolor}         
\usepackage{graphicx}

\usepackage{multirow}
\usepackage[normalem]{ulem}
\usepackage{amsmath}
\usepackage{amssymb}
\newcommand{\redbold}[1]{\textcolor{red}{\textbf{#1}}}
\newcommand{\blueunder}[1]{\textcolor{blue}{\uline{#1}}}
\setcitestyle{numbers,square,comma}
\usepackage{colortbl}
\usepackage[normalem]{ulem}
\definecolor{mycitecolor}{HTML}{0000FF}
\definecolor{myurlcolor}{HTML}{FF8000}
\hypersetup{
  colorlinks=true,
  citecolor=mycitecolor,
  urlcolor=myurlcolor,
  linkcolor=mycitecolor,
}

\definecolor{oddrowgray}{HTML}{E7E6E6}
\definecolor{tocpink}{HTML}{EA028C}
\definecolor{tocgold}{HTML}{FF8000}
\usepackage{pifont}
\usepackage{subcaption}
\usepackage{wrapfig}

\newcommand{\tocboldentry}[2]{%
\noindent\hyperref[#1]{\textbf{#2}}%
\dotfill%
\hyperref[#1]{\textbf{\pageref*{#1}}}%
}

\newcommand{\tocentry}[2]{%
\noindent\hyperref[#1]{#2}%
\dotfill%
\hyperref[#1]{\pageref*{#1}}%
}
\title{TStruct: Learning Shared Temporal Structures for Long-term Time-Series Forecasting}

\author{%
  \textbf{Yangze Li}\textsuperscript{\normalfont 1}{\normalfont,}
  \textbf{Zhengyang Zhou}\textsuperscript{\normalfont 1,2,\Letter}{\normalfont,}
  \textbf{Qihe Huang}\textsuperscript{\normalfont 1}{\normalfont,}
  \textbf{Pengkun Wang}\textsuperscript{\normalfont 1,2}{\normalfont,}
  \textbf{Binwu Wang}\textsuperscript{\normalfont 1,2}{\normalfont,} \\
  \textbf{Yang Wang}\textsuperscript{\normalfont 1,2,\Letter}
  \\[1em]
  \textsuperscript{\normalfont 1}%
  University of Science and Technology of China (USTC), Hefei, China \\
  \textsuperscript{\normalfont 2}%
  Suzhou Institute for Advanced Research, USTC, Suzhou, China \\
  \texttt{liyangze@mail.ustc.edu.cn}
}

\begin{document}

\maketitle

\begingroup
  \renewcommand{\thefootnote}{}
  \makeatletter
  \renewcommand{\@makefnmark}{}
  \makeatother
  \footnotetext[0]{%
    \Letter\ Yang Wang and Zhengyang Zhou are corresponding authors.%
  }
\endgroup

\begin{abstract}
Long-term time series forecasting requires learning temporal dependencies that remain stable and generalizable under temporal distribution shifts. However, real-world time series are often noisy, non-stationary, and heterogeneous, making deep learning models prone to overfitting incidental temporal interactions in the training data. This raises a question: when observations are difficult to align, could temporal relationships be easier to reuse? To investigate this possibility, we analyze patch-level temporal relationship matrices across commonly used time series datasets, revealing dominant patterns shared across different temporal windows and variables. Comparisons with patch-shuffled and IID Gaussian controls further support the presence of reusable temporal relationship structures.
Motivated by these findings, we propose \textbf{T}emporal \textbf{Struct}ure Network (\textbf{TStruct}), a simple yet effective network that learns reusable temporal dependency structures and adaptively combines them for each sample and variable. Specifically, TStruct directly parameterizes a set of shared temporal dependency matrices and generates normalized combination weights using dynamic sample information and static variable priors. These weights represent a soft allocation of each sample-variable pair over shared temporal dependency structures, enabling the model to generate adaptive temporal dependency matrices under shared structural constraints, rather than fitting incidental interactions in a fully unconstrained manner. Notably, TStruct consistently outperforms several strong baselines across all evaluated datasets, achieving superior accuracy while maintaining high computational efficiency. Our code is available at \url{https://github.com/yangzeli/TStruct}.
\end{abstract}

\section{Introduction}
\label{sec:intro}
Long-term time series forecasting (LTSF) has widespread applications in real-world scenarios such as energy scheduling, traffic management, weather forecasting, and financial risk control \citep{informer, autoformer, dlinear, patchtst}. The goal of LTSF is to accurately model future dynamics over a long prediction horizon. The challenge lies not only in fitting historical observations but also in learning temporal dependency structures that remain effective across future time distributions. However, real-world time series data often have low signal-to-noise ratios (SNR), non-stationarity, and temporal distribution shifts, and local fluctuations or incidental interactions in the training data may not generalize well for future predictions~\citep{nonstationarity, low_SNR, distribution_shift, characteristic}. Therefore, a central question in LTSF is: how can we identify and model temporal dependency structures that remain stable across shifting temporal distributions?

\begin{figure}[t]
    \centering
    \includegraphics[width=\linewidth]{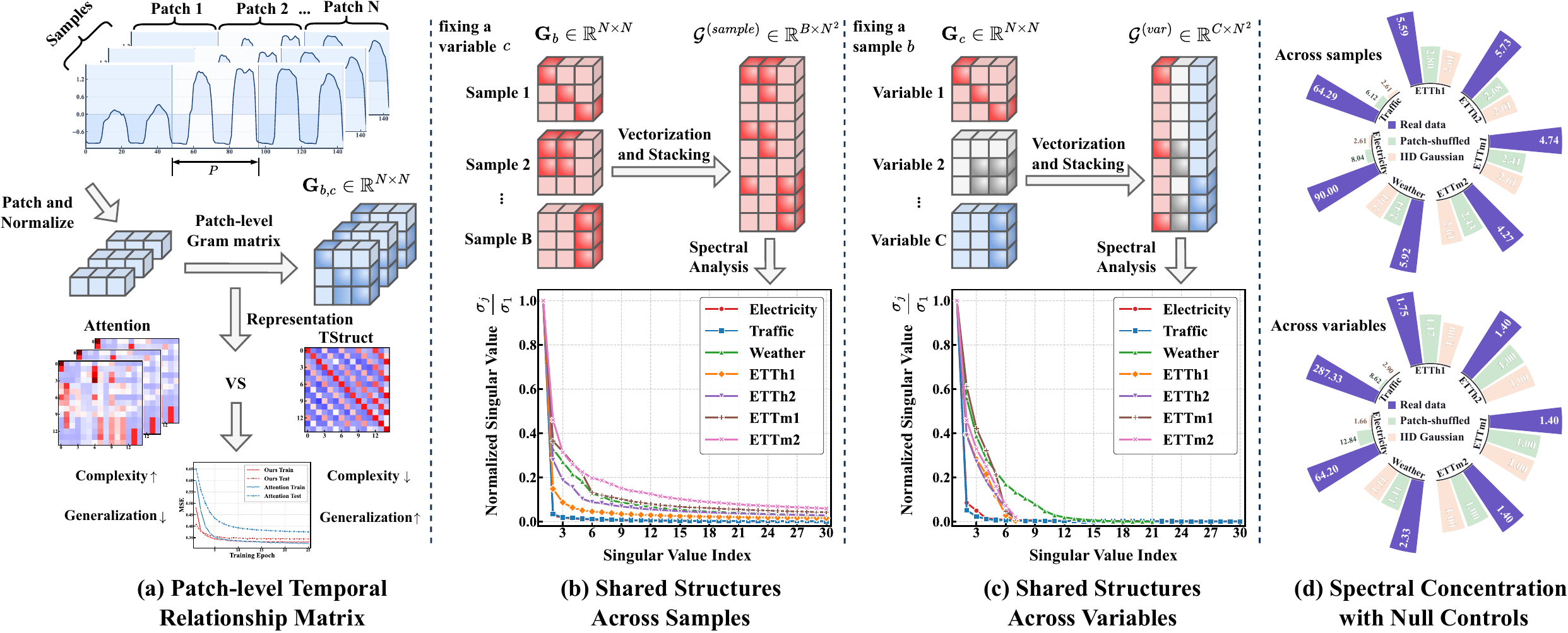}
    \caption{Patch-level temporal relationship structures.
    (a) Construction of patch-level relationship matrices and qualitative comparison between attention-based and TStruct-based temporal dependency modeling.
    (b,c) Concentrated singular value spectra across samples and variables.
    (d) Higher spectral concentration in real data than in patch-shuffled and IID Gaussian controls supports reusable temporal structures.}
    \label{fig:intro}
\end{figure}

In recent years, deep learning methods have driven significant progress in long-term time series forecasting~\citep{informer,autoformer,fedformer,dlinear,patchtst,crossformer,crossgnn,capacity,revisitlinear,chronos,whyattnfails,arellmuseful,timebridge,mafs, phat,storm,spatiotemporal}. One category of methods implicitly models temporal dependencies through model architectures. For example, recurrent models pass historical information via hidden states \citep{segrnn, xlstm}, convolutional and TCN models use shared kernels to extract local or multi-scale temporal patterns \citep{scinet, micn, moderntcn}, Transformer-based models estimate pairwise interactions between time tokens explicitly using attention mechanisms \citep{patchtst, crossformer, pathformer, phaseformer}, while Linear/MLP methods achieve efficient prediction by mixing time or feature dimensions \citep{dlinear, fits, timemixer, sparsetsf, timebase}. Despite their strengths in expressive power and efficiency, these methods still face a trade-off between adaptive temporal dependency modeling and forecasting generalization. Highly flexible interaction mechanisms may fit incidental relationships in low-SNR data, while overly fixed transformations struggle to capture heterogeneous temporal dependencies across different samples and variables \citep{whyattnfails, revisitlinear, characteristic}. A recent line of work directly parameterizes stable temporal patterns, most notably periodicity, through global temporal priors, suggesting that certain temporal structures need not be entirely learned implicitly by generic neural architectures \citep{cyclenet, tqnet, timeemb, mofo, gtr}. However, such explicit parameterization has mainly focused on periodic or time-indexed patterns in the input sequence, leaving open whether temporal dependency structures themselves also contain shared and explicitly learnable patterns.

To investigate whether temporal dependencies contain shared structures, we analyze patch-level temporal relationship matrices across commonly used time series datasets. As shown in Figure~\ref{fig:intro}(a), for each sample-variable sequence, we divide the input window into temporal patches and construct a matrix that captures the relationships between relative patch positions. This representation shifts the analysis from raw value fluctuations to relationship patterns among relative patch positions, allowing us to examine whether these patterns recur across samples and variables. We then collect these matrices separately across samples from different temporal positions and across variables for spectral analysis. Figure~\ref{fig:intro}(b,c) shows clear spectral concentration in both settings. This consistent spectral concentration indicates the presence of shared dominant temporal relationship patterns across samples and variables. Figure~\ref{fig:intro}(d) further examines these patterns using patch-shuffled and IID Gaussian controls under the same analysis pipeline. Independent patch shuffling preserves each matrix's relationship values and spectrum while disrupting the alignment of relative patch positions across samples and variables. The resulting decrease in spectral concentration indicates that shared relationship patterns recur at corresponding relative temporal positions. Compared with IID Gaussian inputs, real data also exhibit stronger concentration, indicating shared regularities beyond the random-input reference. Further details are provided in Appendix~\ref{appendix:dataset_analysis}.

These observations motivate learning reusable temporal structures and adapting their contributions to each sample and variable. We therefore propose the \textbf{T}emporal \textbf{Struct}ure Network (\textbf{TStruct}). Specifically, TStruct directly parameterizes a set of shared temporal dependency matrices and uses dynamic sample information and static variable priors to generate normalized combination weights. These weights serve as a soft allocation of each sample-variable pair over the shared dependency structures, combining the shared matrices into a sample- and variable-specific temporal dependency matrix. In this way, TStruct models heterogeneous temporal dependencies under the constraint of shared structures, balancing generalizable structure and adaptive flexibility. In addition, we introduce the Parameter-efficient Channel Adaptation Head (PCAHead), which utilizes a shared low-rank weight generation mechanism to dynamically produce specific prediction heads for each variable.

The contributions of this paper are summarized as follows:
\begin{itemize}
    \item We reveal that patch-level temporal relationship matrices in real world time series contain dominant structures shared across samples from different temporal positions and across variables.
    \item We propose TStruct, a concise and effective shared-adaptive temporal dependency modeling method that learns shared temporal dependency matrices and uses dynamic sample information and static variable priors to construct sample- and variable-specific dependencies.
    \item We conduct extensive experiments to validate the effectiveness of TStruct. The results demonstrate that this simple design consistently outperforms several strong baselines on standard long-term forecasting benchmarks and remains effective under diverse generalization and robustness evaluations, while also exhibiting high computational efficiency.
\end{itemize}

\section{Related Work}
\label{sec:related_work}

\subsection{Temporal Dependency Modeling for Time Series Forecasting} 
\label{sec:re_1}
Temporal dependency modeling aims to characterize how information interacts across different temporal positions, and is a central problem in long-term time series forecasting. In recent years, Transformer-based and Linear/MLP-based methods have received increasing attention due to their strong performance in long-range dependency modeling and efficient forecasting. Transformer-based methods, such as Informer, Autoformer, PatchTST, Crossformer, and Pathformer \citep{informer,autoformer,patchtst,crossformer,pathformer}, usually use attention mechanisms to dynamically estimate pairwise interactions among temporal tokens or patches for each input, thereby modeling long-range temporal dependencies. Their highly flexible input-dependent interactions provide strong adaptivity, but may also capture sample-specific incidental relations in low-SNR time series. Linear/MLP-based methods, such as DLinear, TSMixer, FiTS, SparseTSF, and TimeBase \citep{dlinear,tsmixer,fits,sparsetsf,timebase}, achieve simple and efficient forecasting through shared temporal or feature-wise mixing. These methods demonstrate the strong competitiveness of fixed shared transformations in LTSF, but their adaptability to heterogeneous temporal dependencies remains limited. In contrast, TStruct neither relies on a single fixed mixing rule nor freely estimates interactions for each input. Instead, it generates normalized information aggregation weights through shared temporal dependency structures and sample-variable-conditioned weights, enabling adaptive temporal dependency modeling under shared structural constraints.

\subsection{Explicit Temporal Structure Parameterization} 
\label{sec:re_2}

Recent studies attempt to explicitly parameterize stable structures in time series, rather than relying entirely on generic neural networks to learn them implicitly from data. CycleNet, TQNet, TimeEmb, MOFO, and GTR \citep{cyclenet,tqnet,timeemb,mofo,gtr} mostly model recurring periodic patterns in data through global learnable periodic representations. These studies suggest that directly parameterizing reusable temporal structures can help improve the generalization ability of time series forecasting. However, existing methods mainly focus on periodic patterns in the input sequence or temporal representations. Temporal dependency structures can contain periodicity, but are not equivalent to periodicity itself; they may also characterize more general relationships among relative temporal positions and patch-level interaction patterns \citep{ts_analysis,patchtst,timesnet}. Unlike these methods, TStruct shifts the explicit parameterization target from input-level periodic patterns to temporal dependency structures themselves.

\section{Methodology}
\label{sec:Methodology}

\textbf{Problem Definition.}
The problem of LTSF can be formulated as follows. Let $\mathbf{X} = \{\mathbf{x}_{t-L+1}, \dots, \mathbf{x}_t\} \in \mathbb{R}^{L \times C}$ denote the sequence of historical observations, where $L$ represents the lookback window length and $C$ denotes the number of variables. At each time step $t$, the observation vector is defined as $\mathbf{x}_t = \{x_t^{(1)}, \dots, x_t^{(C)}\}$, where $x_t^{(c)}$ represents the value of the $c$-th variable. Our objective is to learn a mapping function $\mathcal{F}_\theta$ parameterized by $\theta$ to predict the future values for the next $H$ time steps, denoted as $\mathbf{Y} = \{\mathbf{x}_{t+1}, \dots, \mathbf{x}_{t+H}\} \in \mathbb{R}^{H \times C}$. In this work, we focus on how to learn temporal dependency structures within $\mathcal{F}_{\theta}$ that remain stable and generalizable across different samples and variables.

\subsection{Overview}
\label{subsec:overview}
\begin{figure*}[htb]
    \centering
    \includegraphics[width=\linewidth]{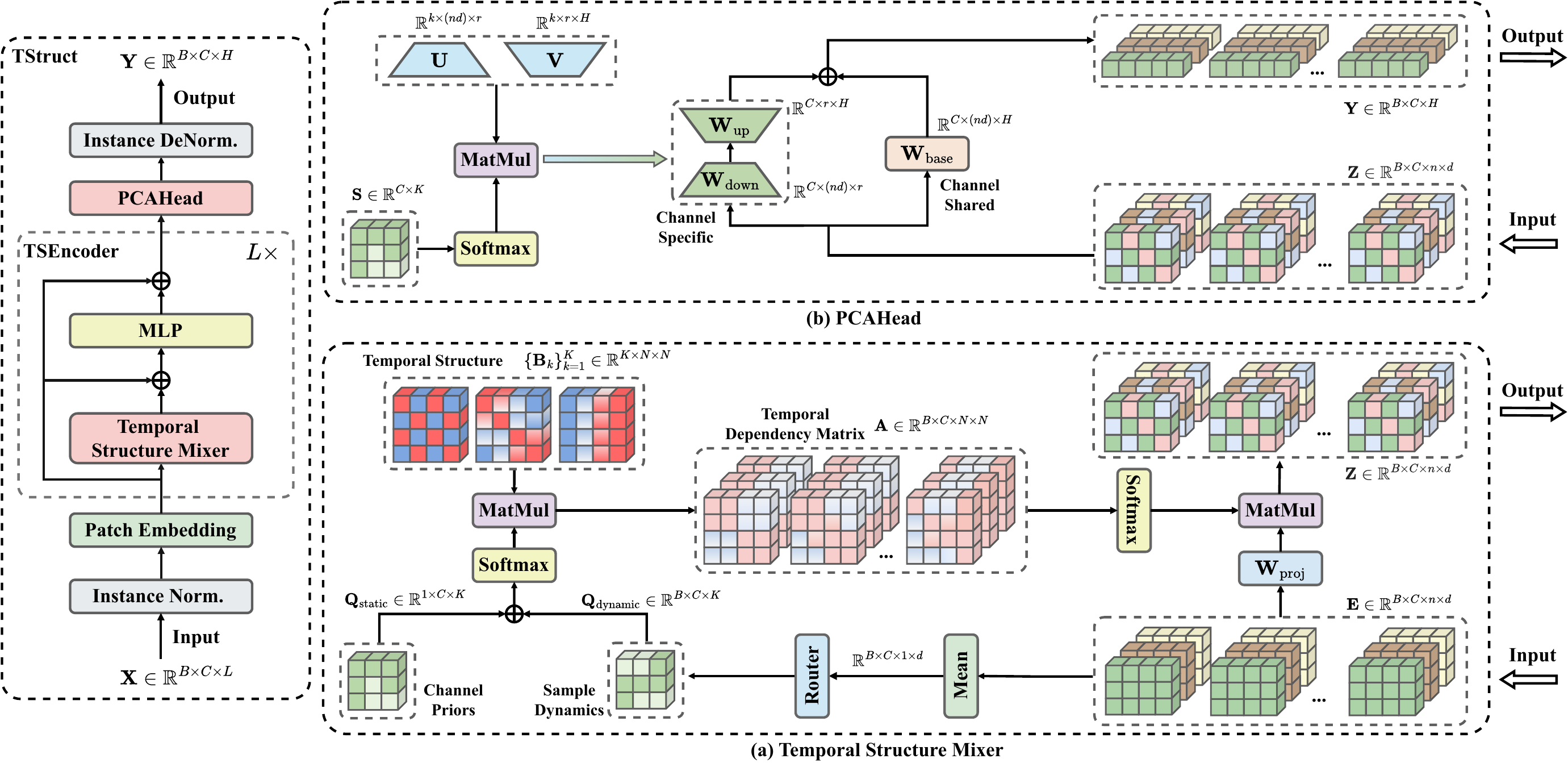}
    \caption{Overall architecture of TStruct. (a) The Temporal Structure Mixer (TSM) generates sample- and variable-specific temporal dependencies by adaptively combining shared temporal dependency structures with dynamic sample information and static variable priors. (b) The PCAHead handles variable heterogeneity by generating variable-aware prediction weights in a parameter-efficient manner.}
    \label{fig:overview}
\end{figure*}

As shown in Figure~\ref{fig:overview}, TStruct contains three components. Following RevIN \citep{revin}, TStruct applies instance-wise normalization before encoding and restores the original scale after prediction. Patch Embedding first converts each univariate sequence into patch-level representations, enabling local pattern extraction and reducing the length of temporal modeling \citep{patchtst,srsnet}. The Temporal Structure Encoder (TSEncoder) then learns shared temporal dependency structures and adaptively combines them using dynamic sample information and static variable priors. The resulting dependency scores are normalized into patch-level aggregation weights, allowing the model to capture sample- and variable-specific temporal dependencies under shared structural constraints. Finally, PCAHead provides a parameter-efficient prediction head that accounts for variable heterogeneity by generating variable-aware forecasting projections.

\subsection{Patch Embedding}
\label{subsec:patch_embedding}
Patch-based representation has been widely used to capture local temporal patterns and reduce the effective sequence length \citep{patchtst}. Given the input sequence of the $c$-th variable $\mathbf{x}^{(c)} \in \mathbb{R}^{L}$, we first segment it into $N$ non-overlapping patches $\mathbf{x}_{\mathrm{p}}^{(c)} \in \mathbb{R}^{N \times P}$, where $P$ is the patch length and $N=L/P$ is the number of patches. For simplicity, we assume that $L$ is divisible by $P$. Subsequently, a learnable linear projection $\mathbf{W}_{\mathrm{emb}} \in \mathbb{R}^{P \times D}$, shared across all variables, maps these patches into a $D$-dimensional latent representation:
\begin{equation}
    \mathbf{x}_{\mathrm{p}}^{(c)} = \operatorname{Unfold}(\mathbf{x}^{(c)}), \quad 
    \mathbf{E}^{(c)} = \mathbf{x}_{\mathrm{p}}^{(c)} \mathbf{W}_{\mathrm{emb}},
\end{equation}
where $\mathbf{E}^{(c)} \in \mathbb{R}^{N \times D}$ denotes the patch-level representation of the $c$-th variable. This representation provides the basis for TSEncoder to model temporal dependencies among relative patch positions.

\subsection{Temporal Structure Encoder}
\label{subsec:tsencoder}
The TSEncoder is designed to model patch-level temporal dependencies in a shared yet adaptive manner. Given the patch-level representation $\mathbf{E} \in \mathbb{R}^{B \times C \times N \times D}$, where $B$, $C$, $N$, and $D$ denote the batch size, number of variables, number of patches, and hidden dimension, respectively, TSEncoder learns temporal dependencies among the $N$ relative patch positions.
We first parameterize a set of globally shared temporal dependency structures:
\begin{equation}
    \mathcal{B}=\{\mathbf{B}_k\}_{k=1}^{K}, 
    \quad \mathbf{B}_k \in \mathbb{R}^{N \times N},
\end{equation}
where $K$ is the number of shared temporal structures. Each $\mathbf{B}_k$ represents a learnable patch-level dependency score pattern.

We initialize the vectorized structure matrices orthogonally, providing distinct initial directions with unit Frobenius norm. This controls the initial scale of the dependency scores before softmax normalization. The importance of initialization is also consistent with findings on learning dynamics in deep linear
networks~\citep{rhee-2026-deep-linear-nets}. A detailed initialization study is provided in Appendix~\ref{appendix:Initialization}.

These shared structures are then combined using sample-variable-specific weights. Specifically, we introduce a learnable static variable prior $\mathbf{Q}_{\mathrm{static}} \in \mathbb{R}^{C \times K}$, which captures variable-specific temporal preferences. A complementary dynamic query is generated by averaging each sample-variable representation along the patch dimension and passing the result through a lightweight router:

\begin{equation}
    \mathbf{Q}_{\mathrm{dynamic}} = \operatorname{Router}\left(\operatorname{Mean}_{N}(\mathbf{E})\right),
\end{equation}
where $\mathbf{Q}_{\mathrm{dynamic}} \in \mathbb{R}^{B \times C \times K}$. The static and dynamic queries are then fused to obtain normalized combination coefficients:
\begin{equation}
    \boldsymbol{\alpha}_{b,c}
    =
    \operatorname{Softmax}\left(
    \mathbf{Q}_{\mathrm{static}}^{(c)} + \mathbf{Q}_{\mathrm{dynamic}}^{(b,c)}
    \right),
\end{equation}
where $\boldsymbol{\alpha}_{b,c} \in \mathbb{R}^{K}$ and the softmax is applied over the $K$ shared structures. The coefficients are nonnegative and sum to one, so the resulting dependency score matrix is a convex combination of the shared bases.

For each sample-variable pair, we compose a temporal dependency score matrix from the shared structures:
\begin{equation}
    \mathbf{S}_{b,c}
    =
    \sum_{k=1}^{K}
    \alpha_{b,c,k} \mathbf{B}_k,
    \quad
    \mathbf{S}_{b,c} \in \mathbb{R}^{N \times N}.
\end{equation}

The composed scores are converted into normalized temporal aggregation weights:
\begin{equation}
    \mathbf{A}_{b,c}
    =
    \operatorname{Softmax}_{\mathrm{row}}
    \left(\mathbf{S}_{b,c}\right).
\end{equation}
Each row of $\mathbf{A}_{b,c}$ assigns nonnegative weights summing to one to the source patches. Temporal aggregation therefore forms weighted averages of projected historical features, constraining the operator to select and combine information without arbitrary signed amplification.

Finally, the normalized temporal dependency matrix is used to aggregate patch-level features:
\begin{equation}
    \widetilde{\mathbf{E}}_{b,c}
    =
    \mathbf{A}_{b,c}
    \left(\mathbf{E}_{b,c}\mathbf{W}_{\mathrm{proj}}\right),
\end{equation}
where $\mathbf{W}_{\mathrm{proj}} \in \mathbb{R}^{D \times D}$ is a learnable feature projection. The aggregated representation is further processed by a feed-forward network with residual connections:
\begin{equation}
    \mathbf{Z}_{b,c}
    =
    \widetilde{\mathbf{E}}_{b,c}
    +
    \operatorname{MLP}\left(\widetilde{\mathbf{E}}_{b,c}\right).
\end{equation}

Together, the shared bases and normalized aggregation weights restrict temporal mixing to a common family of dependency patterns learned across samples and variables. This provides an inductive bias toward stable, reusable temporal structures, while allowing their contributions to be adjusted to the input context.

\subsection{Parameter-efficient Channel Adaptation Head}
\label{sec:pcahead}
PCAHead maps the encoded features to the prediction horizon while balancing variable heterogeneity and parameter efficiency. A fully shared prediction head is efficient but may be insufficient for variable-specific forecasting patterns, whereas assigning an independent head to each variable leads to a parameter cost proportional to $C$. PCAHead therefore adopts a shared base projection with variable-specific low-rank residual adaptation.

Given the encoder output, we flatten the patch and feature dimensions into
$\mathbf{Z}\in\mathbb{R}^{B\times C\times D_{\mathrm{in}}}$, where $D_{\mathrm{in}}=ND$. A shared base projection gives
\begin{equation}
    \mathbf{Y}_{\mathrm{base}}=\mathbf{Z}\mathbf{W}_{\mathrm{base}},
    \quad
    \mathbf{W}_{\mathrm{base}}\in\mathbb{R}^{D_{\mathrm{in}}\times H}.
\end{equation}
For variable-specific adaptation, we define shared low-rank projection banks
$\mathbf{U}_k\in\mathbb{R}^{D_{\mathrm{in}}\times r}$ and
$\mathbf{V}_k\in\mathbb{R}^{r\times H}$, where $r\ll D_{\mathrm{in}}$. For each variable $c$, PCAHead learns normalized coefficients
$\boldsymbol{\alpha}_c=\operatorname{Softmax}(\mathbf{s}^{(c)})$ and assembles
\begin{equation}
    \mathbf{W}_{\mathrm{down}}^{(c)} = \sum_{k=1}^{K} \alpha_{c,k}\mathbf{U}_k,
    \quad
    \mathbf{W}_{\mathrm{up}}^{(c)} = \sum_{k=1}^{K} \alpha_{c,k}\mathbf{V}_k.
\end{equation}
The final prediction is
\begin{equation}
    \hat{\mathbf{Y}}^{(c)}
    =
    \mathbf{Y}_{\mathrm{base}}^{(c)}
    +
    \mathbf{Z}^{(c)}
    \mathbf{W}_{\mathrm{down}}^{(c)}
    \mathbf{W}_{\mathrm{up}}^{(c)}.
\end{equation}
This reduces the head complexity from $O(CD_{\mathrm{in}}H)$ to
$O(D_{\mathrm{in}}H + Kr(D_{\mathrm{in}}+H) + CK)$.
We zero-initialize the up-projection bank so that the residual branch starts from zero, allowing the model to first rely on the shared base prediction and then gradually learn variable-specific corrections.

\section{Experiment}
\label{sec:exp}

\subsection{Experiment Setup}
\label{sec:exp_setup}
Following prior studies, we use Mean Squared Error (MSE) and Mean Absolute Error (MAE) as evaluation metrics. All methods are evaluated under prediction lengths $H=\{96, 192, 336, 720\}$. The datasets and baseline models used for evaluation are summarized below. Dataset statistics are provided in Appendix~\ref{appendix:Dataset_details}, while implementation details and hyperparameter settings are provided in Appendix~\ref{appendix:implementation_details}.

\textbf{Datasets.}
We evaluate TStruct on seven widely used long-term forecasting benchmarks, including Weather, Traffic, Electricity~\citep{timesnet}, and four ETT subsets~\citep{informer}. These datasets cover different domains, variable scales, and temporal patterns, providing a comprehensive evaluation of forecasting performance and generalization ability. 

\textbf{Baselines.}
We compare TStruct with eight strong and recent forecasting baselines, covering different temporal dependency modeling paradigms. Specifically, PatchTST~\citep{patchtst} and PhaseFormer~\citep{phaseformer} represent Transformer-based methods; DLinear~\citep{dlinear} represents simple linear forecasting methods; CycleNet~\citep{cyclenet}, TQNet~\citep{tqnet}, TimeEmb~\citep{timeemb}, and GTR~\citep{gtr} explicitly model temporal patterns or periodic priors; and DUET~\citep{duet} is included as a recent strong forecasting baseline. For all baselines, we report the best results reported in their papers or accompanying code repositories, with the corresponding experimental configurations detailed in Appendix~\ref{appendix:implementation_details}.

\subsection{Main results}
\label{sec:main_results}

\definecolor{yearblue}{rgb}{0.0, 0.0, 0.5}
\definecolor{bestredbg}{HTML}{FFCCCC}
\definecolor{secondbluebg}{HTML}{CCE5FF}
\definecolor{methodheaderbg}{HTML}{D6DCE4}
\definecolor{metricheaderbg}{HTML}{EFEFEF}

\providecommand{\redbold}[1]{}
\renewcommand{\redbold}[1]{\cellcolor{bestredbg}\textcolor{red}{\textbf{#1}}}
\providecommand{\blueunder}[1]{}
\renewcommand{\blueunder}[1]{\cellcolor{secondbluebg}\textcolor{blue}{\uline{#1}}}

\begin{table*}[htbp]
  \centering
  \caption{Average multivariate time series forecasting results over all prediction horizons. The best results are marked in \textcolor{red}{\textbf{bold red}}, and the second-best results are marked in \textcolor{blue}{\uline{blue underline}}.}
  \label{tab:summary_results}
  \resizebox{\linewidth}{!}{
    \setlength{\tabcolsep}{3pt}
    \renewcommand{\arraystretch}{1.1}
    \begin{tabular}{c|cc|cc|cc|cc|cc|cc|cc|cc|cc}
    \toprule[1.1pt]
    \rowcolor{methodheaderbg}
    \multicolumn{1}{>{\columncolor{methodheaderbg}}c|}{\raisebox{0.5\normalbaselineskip}{\textbf{Methods}}} &
    \multicolumn{2}{>{\columncolor{methodheaderbg}}c|}{\shortstack[c]{\textbf{TStruct}\\ (ours)}} &
    \multicolumn{2}{>{\columncolor{methodheaderbg}}c|}{\shortstack[c]{\textbf{DUET}\\ \citeyearpar{duet}}} &
    \multicolumn{2}{>{\columncolor{methodheaderbg}}c|}{\shortstack[c]{\textbf{PhaseFormer}\\ \citeyearpar{phaseformer}}} &
    \multicolumn{2}{>{\columncolor{methodheaderbg}}c|}{\shortstack[c]{\textbf{GTR}\\ \citeyearpar{gtr}}} &
    \multicolumn{2}{>{\columncolor{methodheaderbg}}c|}{\shortstack[c]{\textbf{TimeEmb}\\ \citeyearpar{timeemb}}} &
    \multicolumn{2}{>{\columncolor{methodheaderbg}}c|}{\shortstack[c]{\textbf{TQNet}\\ \citeyearpar{tqnet}}} &
    \multicolumn{2}{>{\columncolor{methodheaderbg}}c|}{\shortstack[c]{\textbf{CycleNet}\\ \citeyearpar{cyclenet}}} &
    \multicolumn{2}{>{\columncolor{methodheaderbg}}c|}{\shortstack[c]{\textbf{PatchTST}\\ \citeyearpar{patchtst}}} &
    \multicolumn{2}{>{\columncolor{methodheaderbg}}c}{\shortstack[c]{\textbf{DLinear}\\ \citeyearpar{dlinear}}} \\
    \cmidrule[0.9pt](r){1-1}\cmidrule[0.9pt](lr){2-3}\cmidrule[0.9pt](lr){4-5}\cmidrule[0.9pt](lr){6-7}\cmidrule[0.9pt](lr){8-9}\cmidrule[0.9pt](lr){10-11}\cmidrule[0.9pt](lr){12-13}\cmidrule[0.9pt](lr){14-15}\cmidrule[0.9pt](lr){16-17}\cmidrule[0.9pt](l){18-19}
    \rowcolor{metricheaderbg}
    \textbf{Datasets} & \textbf{MSE} & \textbf{MAE} & \textbf{MSE} & \textbf{MAE} & \textbf{MSE} & \textbf{MAE} & \textbf{MSE} & \textbf{MAE} & \textbf{MSE} & \textbf{MAE} & \textbf{MSE} & \textbf{MAE} & \textbf{MSE} & \textbf{MAE} & \textbf{MSE} & \textbf{MAE} & \textbf{MSE} & \textbf{MAE} \\
    \midrule[1.1pt]
    ETTh1 & \redbold{0.390} & \redbold{0.415} & \blueunder{0.398} & 0.419 & 0.403 & \redbold{0.415} & 0.415 & 0.419 & 0.418 & 0.433 & 0.420 & 0.428 & 0.415 & 0.426 & 0.417 & 0.431 & 0.425 & 0.439 \\
    \midrule
    ETTh2 & \redbold{0.331} & \redbold{0.381} & \blueunder{0.334} & \blueunder{0.383} & 0.347 & 0.389 & 0.346 & 0.394 & 0.353 & 0.396 & 0.358 & 0.399 & 0.345 & 0.394 & 0.351 & 0.395 & 0.470 & 0.468 \\
    \midrule
    ETTm1 & \redbold{0.337} & \redbold{0.367} & \blueunder{0.338} & \blueunder{0.369} & 0.347 & 0.374 & 0.345 & 0.377 & 0.345 & 0.376 & 0.353 & 0.388 & 0.352 & 0.378 & 0.349 & 0.381 & 0.356 & 0.378 \\
    \midrule
    ETTm2 & \redbold{0.243} & \redbold{0.304} & \blueunder{0.248} & \blueunder{0.307} & 0.251 & 0.314 & 0.255 & 0.315 & \blueunder{0.248} & 0.308 & 0.258 & 0.318 & 0.249 & 0.311 & 0.256 & 0.314 & 0.259 & 0.324 \\
    \midrule
    Weather & \redbold{0.213} & \redbold{0.249} & \blueunder{0.218} & \blueunder{0.252} & 0.223 & 0.261 & 0.227 & 0.264 & \blueunder{0.218} & 0.257 & 0.230 & 0.266 & 0.237 & 0.277 & 0.224 & 0.262 & 0.242 & 0.293 \\
    \midrule
    Electricity & \redbold{0.151} & \redbold{0.244} & \blueunder{0.157} & \blueunder{0.248} & 0.161 & 0.250 & 0.160 & 0.254 & 0.158 & 0.253 & 0.162 & 0.256 & 0.157 & 0.250 & 0.162 & 0.254 & 0.167 & 0.264 \\
    \midrule
    Traffic & \redbold{0.381} & \blueunder{0.252} & 0.393 & 0.256 & \blueunder{0.387} & \redbold{0.250} & 0.397 & 0.274 & 0.399 & 0.272 & 0.404 & 0.282 & 0.406 & 0.280 & 0.396 & 0.266 & 0.418 & 0.287 \\
    \midrule
    \multicolumn{1}{c|}{\textbf{1st Count}} & \redbold{7} & \redbold{6} & \blueunder{0} & \blueunder{0} & \textbf{0} & \textbf{2} & \textbf{0} & \textbf{0} & \textbf{0} & \textbf{0} & \textbf{0} & \textbf{0} & \textbf{0} & \textbf{0} & \textbf{0} & \textbf{0} & \textbf{0} & \textbf{0} \\
    \bottomrule[1.1pt]
    \end{tabular}
  }
\end{table*}

Table~\ref{tab:summary_results} reports the average results over all prediction horizons, with detailed comparisons in Table~\ref{tab:main_results}. TStruct achieves the best average MSE on all seven benchmarks and the best average MAE on six, ranking first in 13 of the 14 dataset-metric comparisons. These results demonstrate the effectiveness of shared-adaptive temporal dependency modeling against strong Transformer-based, Linear-based, and explicit temporal-pattern modeling baselines.

\emph{On ETT and Weather}, TStruct shows clear advantages over existing methods. These datasets often contain non-stationary temporal dynamics and distribution variations, where overly flexible models may fit incidental temporal interactions and overly fixed transformations may be insufficiently adaptive. The consistent improvements on these datasets indicate that TStruct can better capture stable temporal dependency structures while preserving sample-specific adaptability. 

\emph{On Electricity and Traffic}, Notably, without explicit cross-variable interactions, TStruct achieves the best average MSE on both datasets, with the best average MAE on Electricity and the second-best on Traffic. This result is consistent with our spectral analysis in Appendix~\ref{appendix:temporal_structure_analysis}: Electricity and Traffic exhibit stronger spectral concentration in patch-level temporal relationship matrices across both samples and variables, suggesting a higher degree of shared temporal dependency structures. Such structures can be effectively captured by TStruct through shared temporal dependency matrices and sample-variable-conditioned adaptation.

\subsection{Ablation Study}
\label{sec:abl}
We evaluate the effectiveness of three key components in TStruct. The full model is compared against three variants:
\textbf{\ding{172} TSM $\rightarrow$ MHA} replaces the Temporal Structure Mixer with multi-head attention;
\textbf{\ding{173} TSM $\rightarrow$ Linear} replaces it with a shared linear temporal mixer; and
\textbf{\ding{174} PCAHead $\rightarrow$ Linear} replaces PCAHead with a shared linear forecasting head.
All TStruct variants in the ablation study are trained with the same objective and training protocol. 
The summary of these ablation results is reported in Table~\ref{tab:ablation_summary}.
\definecolor{headergray}{HTML}{D6DCE4}
\definecolor{metricgray}{HTML}{EFEFEF}
\definecolor{bestredbg}{HTML}{FFCCCC}

\begin{table*}[t]
\centering
\caption{
Ablation study on the main components of TStruct. 
Results are averaged over all prediction horizons.
Detailed results for each prediction horizon are provided in Table~\ref{tab:ablation_results}.
}
\label{tab:ablation_summary}
\vspace{0.03in}

\footnotesize
\setlength{\tabcolsep}{4.2pt}
\renewcommand{\arraystretch}{1.10}

\resizebox{0.99\textwidth}{!}{
\begin{tabular}{>{\centering\arraybackslash}m{1cm}|cc|cc|cc|cc|cc|cc|cc}
\toprule[1.1pt]
\rowcolor{headergray}
\textbf{Dataset}
& \multicolumn{2}{c|}{\textbf{ETTh1}}
& \multicolumn{2}{c|}{\textbf{ETTh2}}
& \multicolumn{2}{c|}{\textbf{ETTm1}}
& \multicolumn{2}{c|}{\textbf{ETTm2}}
& \multicolumn{2}{c|}{\textbf{Weather}}
& \multicolumn{2}{c|}{\textbf{Electricity}}
& \multicolumn{2}{c}{\textbf{Traffic}} \\
\cmidrule[0.9pt](r){1-1}
\cmidrule[0.9pt](lr){2-3}
\cmidrule[0.9pt](lr){4-5}
\cmidrule[0.9pt](lr){6-7}
\cmidrule[0.9pt](lr){8-9}
\cmidrule[0.9pt](lr){10-11}
\cmidrule[0.9pt](lr){12-13}
\cmidrule[0.9pt](l){14-15}
\rowcolor{metricgray}
\textbf{Metric}
& \textbf{MSE} & \textbf{MAE}
& \textbf{MSE} & \textbf{MAE}
& \textbf{MSE} & \textbf{MAE}
& \textbf{MSE} & \textbf{MAE}
& \textbf{MSE} & \textbf{MAE}
& \textbf{MSE} & \textbf{MAE}
& \textbf{MSE} & \textbf{MAE} \\
\midrule[1.0pt]
\ding{172}
& 0.425 & 0.444
& 0.350 & 0.395
& 0.376 & 0.391
& 0.259 & 0.318
& 0.229 & 0.265
& 0.162 & 0.254
& 0.403 & 0.270 \\
\midrule
\ding{173}
& 0.403 & 0.424
& 0.341 & 0.384
& 0.347 & 0.372
& 0.248 & 0.308
& 0.218 & 0.254
& 0.153 & 0.246
& 0.382 & 0.253 \\
\midrule
\ding{174}
& 0.394 & 0.417
& 0.339 & 0.386
& 0.341 & 0.368
& 0.248 & 0.307
& 0.215 & 0.251
& 0.152 & 0.245
& 0.382 & 0.253 \\
\midrule
\textbf{TStruct}
& \redbold{0.390} & \redbold{0.415}
& \redbold{0.331} & \redbold{0.381}
& \redbold{0.337} & \redbold{0.367}
& \redbold{0.243} & \redbold{0.304}
& \redbold{0.213} & \redbold{0.249}
& \redbold{0.151} & \redbold{0.244}
& \redbold{0.381} & \redbold{0.252} \\
\bottomrule[1.1pt]
\end{tabular}
}
\end{table*}

\emph{On ETT and Weather},
TStruct clearly outperforms both \textbf{\ding{172}} and \textbf{\ding{173}} on these datasets.
These datasets exhibit stronger temporal distribution variations, where forecasting requires both shared structures for generalization and adaptive modeling for sample- and variable-specific heterogeneity.
MHA provides flexible input-dependent interactions but may fit incidental relations, while a shared linear mixer is stable but insufficiently adaptive.
The superior performance of TStruct indicates that shared-adaptive temporal dependency modeling is especially beneficial in such settings.

\emph{On Electricity and Traffic},
The gap between TStruct and \textbf{\ding{173}} becomes smaller on these datasets, while TStruct still clearly outperforms \textbf{\ding{172}}.
This is consistent with our spectral analysis in Appendix~\ref{appendix:dataset_analysis}: Electricity and Traffic exhibit stronger cross-sample and cross-variable spectral concentration in patch-level temporal relationship matrices, suggesting more pronounced shared temporal dependency structures.
Therefore, simple shared temporal mixing is already highly competitive on these datasets, whereas fully input-dependent attention remains less stable.

Finally, replacing PCAHead with a shared linear head leads to moderate but consistent degradation, showing that variable-aware prediction further improves performance under variable heterogeneity.

\subsection{Generalization Analysis}
\label{sec:genera}
\begin{figure}[t]
    \centering
    \begin{subfigure}[t]{0.42\linewidth}
        \centering
        \includegraphics[width=\linewidth]{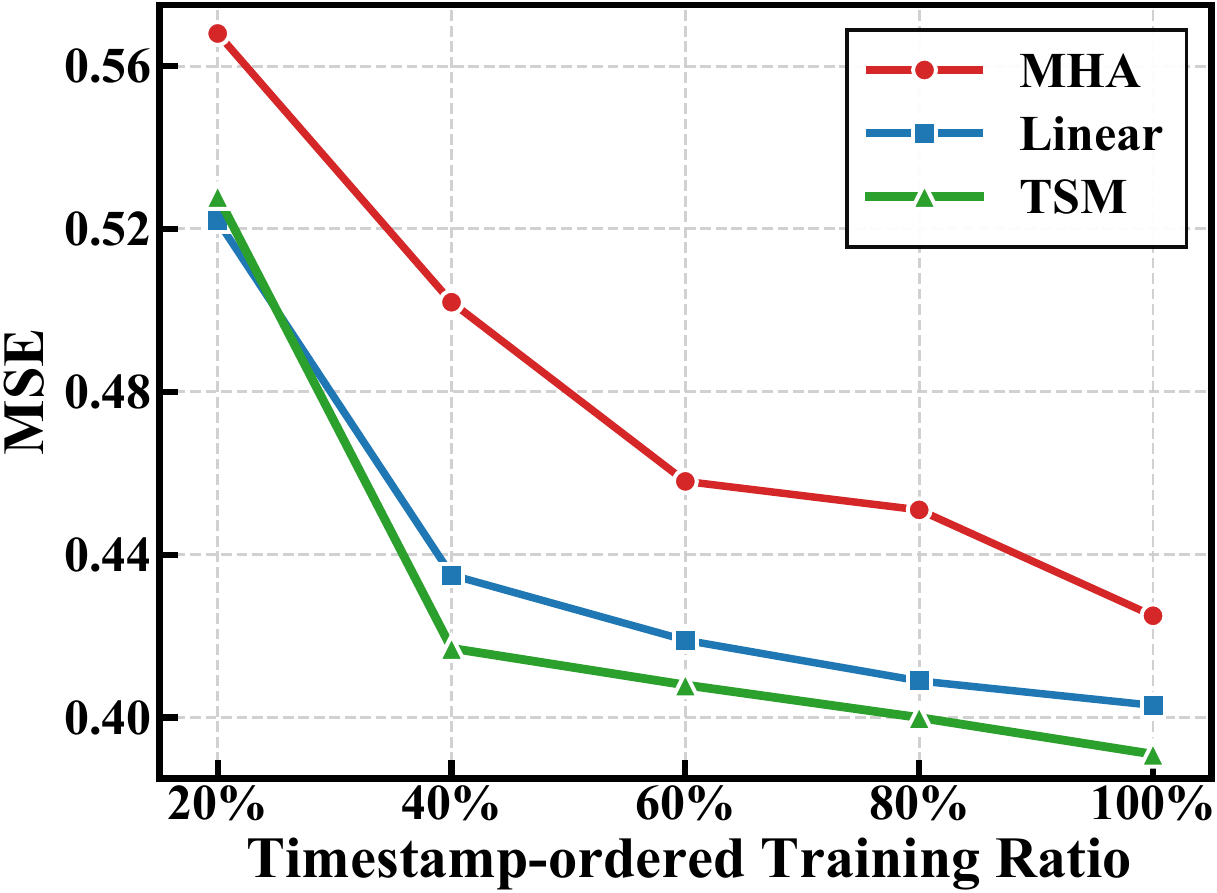}
        \caption{Earliest samples.}
        \label{fig:earliest_samples}
    \end{subfigure}
    \hfill
    \begin{subfigure}[t]{0.42\linewidth}
        \centering
        \includegraphics[width=\linewidth]{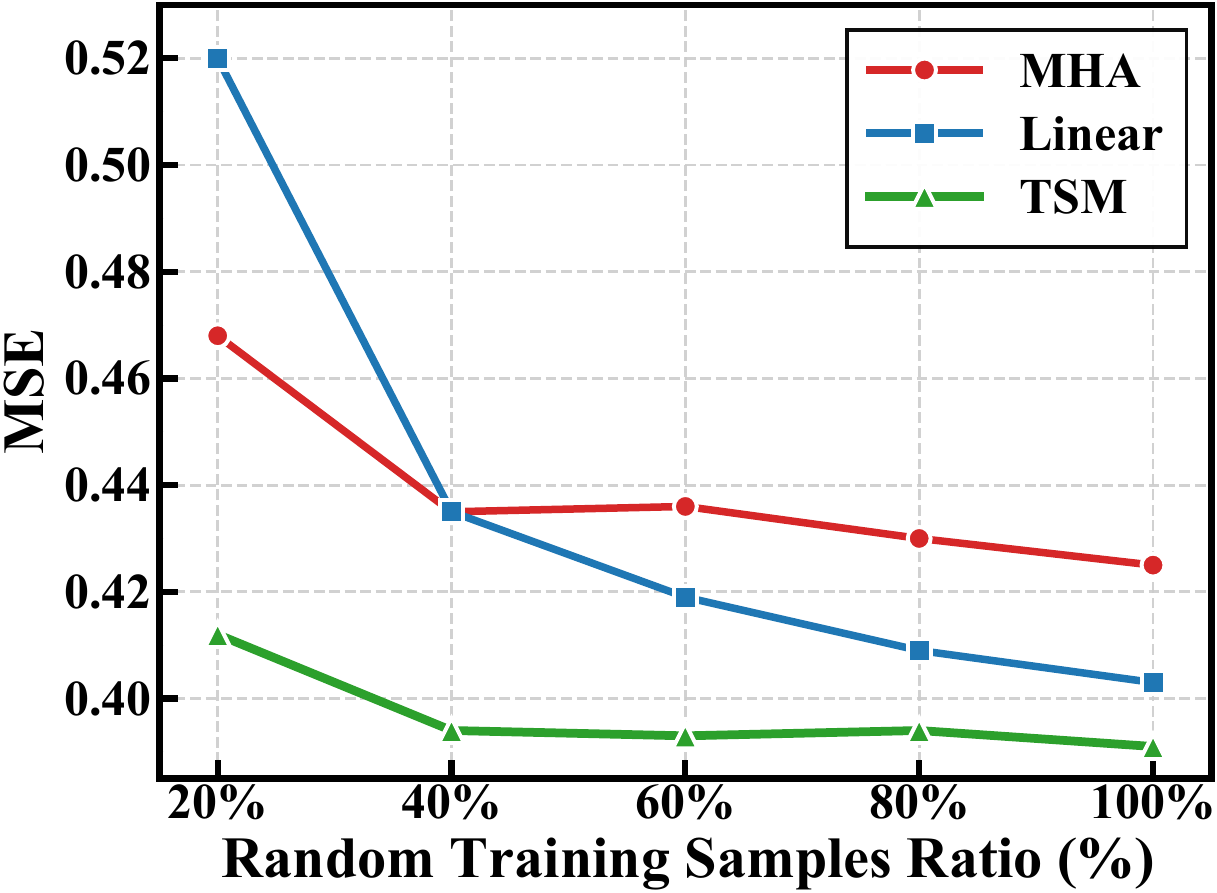}
        \caption{Random samples.}
        \label{fig:random_samples}
    \end{subfigure}
    \caption{
    Generalization analysis on ETTh1.
    (a) Models are trained with the earliest timestamp-ordered training samples, evaluating temporal distribution shift.
    (b) Models are trained with randomly sampled training instances, evaluating sample efficiency.
    }
    \label{fig:generalization_etth1}
\end{figure}
To evaluate generalization, we conduct two reduced-training experiments on ETTh1. We compare the proposed TSM with two variants that replace TSM with Linear or MHA, keeping all other components and experimental settings identical. The earliest sample setting evaluates temporal distribution shift, while random sample setting evaluates sample efficiency.

\emph{Temporal Distribution Shift.}
As shown in Figure~\ref{fig:generalization_etth1}(a), models are trained using the earliest $20\%, 40\%, 60\%, 80\%, 100\%$ of timestamp-ordered training samples, while validation and test sets remain unchanged. 
MHA consistently performs worse than Linear and TSM, indicating that fully input-dependent attention is more sensitive to temporally biased training data. 
TSM is close to Linear when only the earliest $20\%$ samples are used, but clearly outperforms Linear from $40\%$ onward, showing that shared-adaptive temporal dependency modeling improves generalization once sufficient shared structures can be estimated.

\emph{Sample Efficiency.}
As shown in Figure~\ref{fig:generalization_etth1}(b), TSM achieves the best performance even with only $20\%$ randomly sampled training data and quickly converges around $40\%$. 
Notably, TSM trained with $40\%$ data already outperforms MHA and Linear trained with the full training set. 
This suggests that shared temporal structures improve data efficiency by reducing the need to relearn temporal interactions from individual samples.

\subsection{Effect of Look-back Length}
\label{sec:look_back}
\begin{wrapfigure}{r}{0.44\linewidth}
    \centering
    \vspace{-0.35in}
    \includegraphics[width=0.95\linewidth]{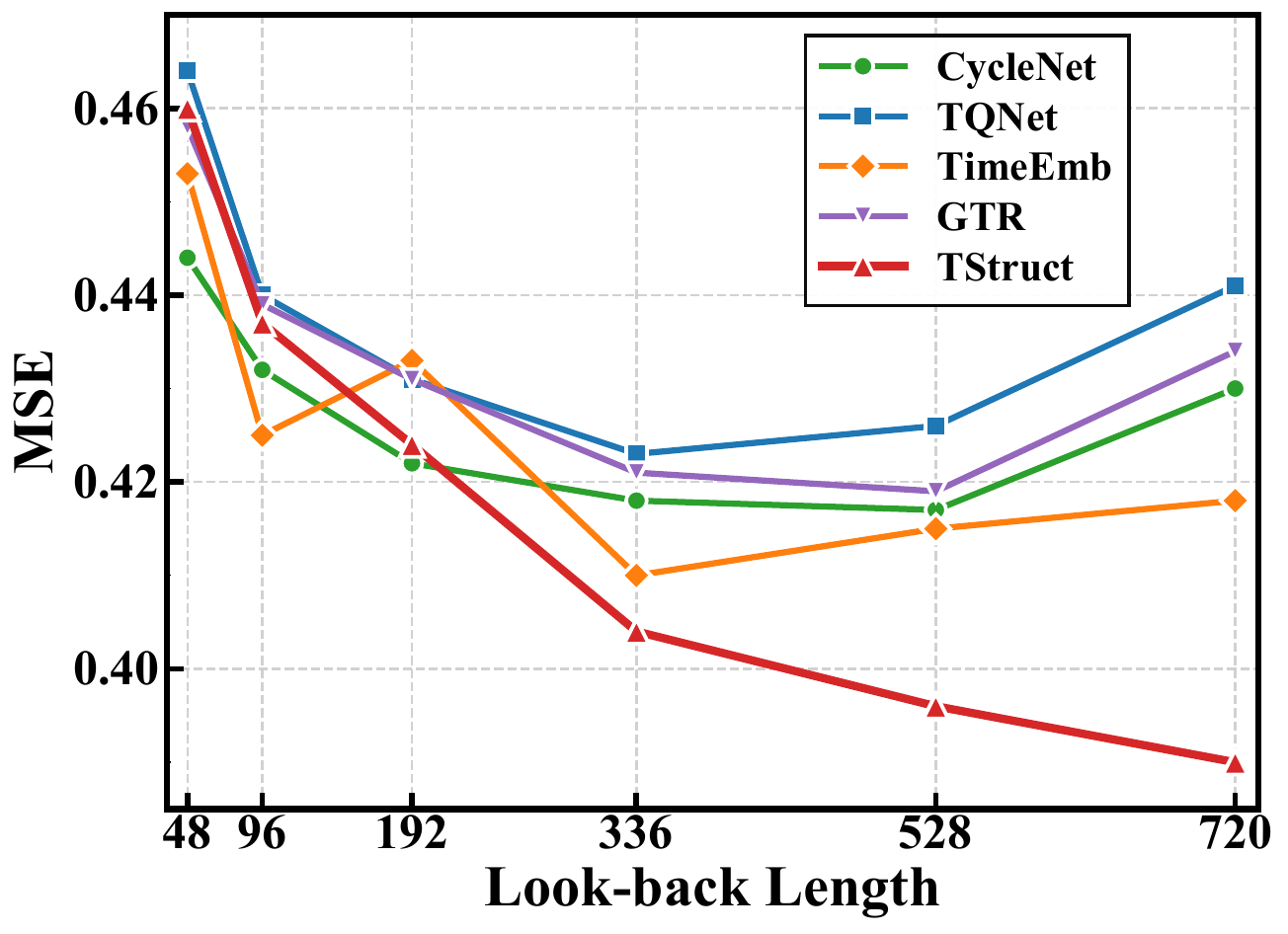}
    \caption{Effect of look-back length on ETTh1.}
    \label{fig:lookback_mse}
    \vspace{-0.3in}
\end{wrapfigure}  

To examine how different methods utilize historical context, we vary the look-back length from $48$ to $720$ and compare TStruct with explicit temporal-pattern parameterization methods, including CycleNet~\citep{cyclenet}, TQNet~\citep{tqnet}, TimeEmb~\citep{timeemb}, and GTR~\citep{gtr}. These baselines mainly rely on global periodic representations, which can remain effective even with short input windows because the periodic priors are largely input-independent. However, as the look-back length increases, their improvements become limited and may even degrade when overly long histories introduce noisy or less relevant observations.

In contrast, TStruct consistently benefits from longer look-back windows. This suggests that TStruct does not merely rely on fixed global periodic patterns; instead, it explicitly models patch-level temporal dependency structures, which become richer as more historical context is available. This further indicates that patch-level temporal dependency structures can provide a more extensible modeling target than fixed global periodic priors when longer historical context is available.

\subsection{Interpretability of Learned Temporal Structures}
\label{sec:Interpretability}
\begin{figure*}[t]
    \centering
    \includegraphics[width=\linewidth]{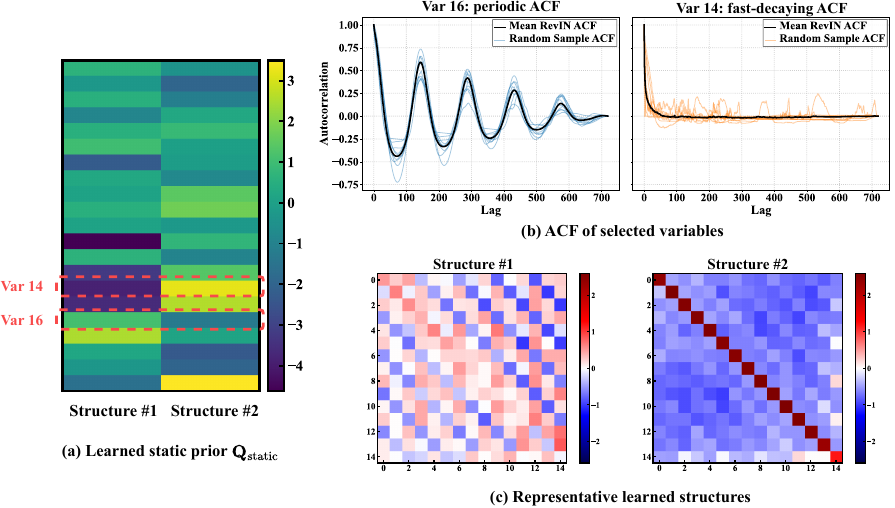} 
    \caption{Interpretability analysis on the Weather dataset.
    Different variables exhibit different ACF patterns and assign different weights to the learned shared temporal structures through $Q_{\mathrm{static}}$. The representative learned structures show distinct patch-level dependency patterns, including broader recurrent relations and local diagonal-dominant continuity.}
    \label{fig:case_analysis}
    \vspace{-0.25in}
\end{figure*}

To examine whether the learned structures are interpretable, we visualize the learned static prior, the autocorrelation functions (ACF) of representative variables, and the corresponding temporal dependency structures on the Weather dataset. As shown in Figure~\ref{fig:case_analysis}, Var 16 exhibits clear periodic autocorrelation, while Var 14 shows rapidly decaying autocorrelation, indicating that these two variables have different temporal characteristics. The learned static prior $\mathbf{Q}_{\mathrm{static}}$ assigns different preferences to these variables over the shared structures. Specifically, Var 16 places higher weight on Structure \#1, which captures broader recurrent relations among patches, whereas Var 14 assigns higher weight to Structure \#2, which shows strong diagonal dominance and reflects local temporal continuity. This suggests that $\mathbf{Q}_{\mathrm{static}}$ captures variable-specific temporal preferences rather than using the same dependency pattern for all variables. Overall, these observations indicate that TStruct learns reusable temporal dependency structures and adapts them to variables with different temporal characteristics through the static prior. For additional visualizations of learned structure matrices on other datasets, please refer to Appendix~\ref{appendix:Interpretability}.

\section{Conclusion}
\label{sec:conclusion}

In this work, we investigated whether temporal relationships can be reused across noisy and heterogeneous time series. Analysis of patch-level relationship matrices reveals dominant patterns shared across temporal windows and variables, with null controls further supporting their reusability. Motivated by these findings, we proposed TStruct, a simple yet effective network that learns shared temporal dependency structures and adaptively combines them for each sample and variable. Experiments on seven forecasting benchmarks show that TStruct consistently outperforms strong baselines while maintaining high computational efficiency. Ablation and generalization analyses further highlight the value of combining shared structures with sample-specific adaptation.

\section*{Acknowledgement}
This paper is partially supported by the National Natural Science Foundation of China (No.62502488, No.12227901), Natural Science Foundation of Jiangsu Province (BK20240460), the grant from State Key Laboratory of Resources and Environmental Information System. The AI-driven experiments, simulations and model training were performed on the robotic AI-Scientist platform of Chinese Academy of Sciences. The authors declare no competing interests.

\bibliography{cited_papers}

\clearpage

\begingroup
\hypersetup{linkcolor=tocgold}
\setlength{\parindent}{0pt}
\setlength{\parskip}{0pt}

{\fontsize{18pt}{30pt}\selectfont\bfseries Appendix\par}
\vspace{14pt}

\phantomsection
{\fontsize{14pt}{24pt}\selectfont\bfseries Catalogue\par}
\label{sec:Catalogue}
\vspace{-3pt}

{\color{black!65}\noindent\rule{\linewidth}{1.1pt}\par}
\vspace{8pt}

\newcommand{\tocsectiongap}{\\[0.65em]}
\newcommand{\tocsubgap}{\\[0.25em]}
\newcommand{\tocheadinggap}{\\[0.25em]}

\noindent\hfill%
\begin{minipage}{0.90\linewidth}
\large
\setlength{\parindent}{0pt}
\setlength{\parskip}{0pt}

\noindent\tocboldentry{sec:intro}{1. Introduction}
\tocsectiongap
\noindent\tocboldentry{sec:related_work}{2. Related Work}
\tocheadinggap
\noindent\hspace*{1.5em}\tocentry{sec:re_1}{2.1 Temporal Dependency Modeling for Time Series Forecasting}
\tocsubgap
\noindent\hspace*{1.5em}\tocentry{sec:re_2}{2.2 Explicit Temporal Structure Parameterization}
\tocsectiongap
\noindent\tocboldentry{sec:Methodology}{3. Methodology}
\tocheadinggap
\noindent\hspace*{1.5em}\tocentry{subsec:overview}{3.1 Overview}
\tocsubgap
\noindent\hspace*{1.5em}\tocentry{subsec:patch_embedding}{3.2 Patch Embedding}
\tocsubgap
\noindent\hspace*{1.5em}\tocentry{subsec:tsencoder}{3.3 Temporal Structure Encoder}
\tocsubgap
\noindent\hspace*{1.5em}\tocentry{sec:pcahead}{3.4 Parameter-efficient Channel Adaptation Head}
\tocsectiongap
\noindent\tocboldentry{sec:exp}{4. Experiment}
\tocheadinggap
\noindent\hspace*{1.5em}\tocentry{sec:exp_setup}{4.1 Experiment Setup}
\tocsubgap
\noindent\hspace*{1.5em}\tocentry{sec:main_results}{4.2 Main Results}
\tocsubgap
\noindent\hspace*{1.5em}\tocentry{sec:abl}{4.3 Ablation Study}
\tocsubgap
\noindent\hspace*{1.5em}\tocentry{sec:genera}{4.4 Generalization Analysis}
\tocsubgap
\noindent\hspace*{1.5em}\tocentry{sec:look_back}{4.5 Effect of Look-back Length}
\tocsubgap
\noindent\hspace*{1.5em}\tocentry{sec:Interpretability}{4.6 Interpretability of Learned Temporal Structures}
\tocsectiongap
\noindent\tocboldentry{sec:conclusion}{5. Conclusion}
\tocsectiongap
\noindent\tocboldentry{appendix:dataset_analysis}{Appendix}
\tocheadinggap
\noindent\hspace*{1.5em}\tocentry{appendix:dataset_analysis}{A. Dataset and Temporal Structure Analysis}
\tocsubgap
\noindent\hspace*{1.5em}\tocentry{appendix:implementation_details}{B. Implementation Details}
\tocsubgap
\noindent\hspace*{1.5em}\tocentry{appendix:robust_sensitivity}{C. Additional Experimental Analyses}
\tocsubgap
\noindent\hspace*{1.5em}\tocentry{appendix:Interpretability}{D. Additional Interpretability Visualizations}
\tocsubgap
\noindent\hspace*{1.5em}\tocentry{appendix:detailed_results}{E. Detailed Experimental Results}
\tocsubgap
\noindent\hspace*{1.5em}\tocentry{appendix:limitations}{F. Limitations and Future Directions}
\par

\end{minipage}\hfill\null
\par

\vspace{8pt}
{\color{black!65}\noindent\rule{\linewidth}{1.1pt}\par}

\endgroup

\clearpage

\appendix

\section{Dataset and Temporal Structure Analysis}
\label{appendix:dataset_analysis}

\subsection{Dataset Details}
\label{appendix:Dataset_details}
We evaluate TStruct on seven widely used long-term forecasting benchmarks, including four ETT subsets, Weather, Electricity, and Traffic. The statistics of these datasets are summarized in Table~\ref{dataset_description}.
The ETT datasets~\citep{informer} contain temperature and power load records collected from electricity transformers in two regions of China from 2016 to 2018. They include two temporal granularities: hourly records, denoted as ETTh1 and ETTh2, and 15-minute records, denoted as ETTm1 and ETTm2. 
The Weather dataset~\citep{timesnet} contains 21 meteorological variables recorded at 10-minute intervals in 2020, including air temperature, humidity, visibility, and other weather-related indicators. 
The Electricity dataset~\citep{timesnet} consists of hourly electricity consumption records from 321 clients, measured in kilowatt-hours (kWh), and is derived from the UCI Machine Learning Repository. 
The Traffic dataset~\citep{timesnet} records hourly road occupancy rates collected by 862 sensors on freeways in the San Francisco Bay Area, reflecting large-scale traffic flow dynamics.

\begin{table}[htbp]
  \caption{Detailed descriptions of datasets.}
  \label{dataset_description}
  \centering
  \small
  \setlength{\tabcolsep}{4pt}
  \renewcommand{\arraystretch}{1.05}
  \definecolor{oddrowgray}{HTML}{E7E6E6}

  \begin{tabular}{c|cccccc}
    \toprule
    \textbf{Dataset} & \textbf{Variables} & \textbf{Timestamps} & \textbf{Split} & \textbf{Prediction Length} & \textbf{Sampling} & \textbf{Information} \\
    \midrule
    \rowcolor{oddrowgray}
    ETTh1   & 7   & 14400 & 6:2:2 & \{96,192,336,720\} & Hourly & Electricity \\
    ETTh2   & 7   & 14400 & 6:2:2 & \{96,192,336,720\} & Hourly & Electricity \\
    \rowcolor{oddrowgray}
    ETTm1   & 7   & 57600 & 6:2:2 & \{96,192,336,720\} & 15min  & Electricity \\
    ETTm2   & 7   & 57600 & 6:2:2 & \{96,192,336,720\} & 15min  & Electricity \\
    \rowcolor{oddrowgray}
    Weather & 21  & 52696 & 7:1:2 & \{96,192,336,720\} & 10min  & Weather \\
    ECL     & 321 & 26304 & 7:1:2 & \{96,192,336,720\} & Hourly & Electricity \\
    \rowcolor{oddrowgray}
    Traffic & 862 & 17544 & 7:1:2 & \{96,192,336,720\} & Hourly & Transport \\
    \bottomrule
  \end{tabular}
\end{table}

\subsection{Temporal Structure Analysis}
\label{appendix:temporal_structure_analysis}

\paragraph{Patch-level temporal relationship matrix.} Real-world time series often have low-SNR, and temporal interactions observed in individual windows may reflect incidental fluctuations. We hypothesize that these observations nevertheless contain relatively stable temporal structures associated with the underlying evolution, which can be extracted from recurring relationships across windows and variables. To examine this hypothesis, we construct patch-level temporal relationship matrices that describe relationships among relative temporal positions. For each dataset, we first standardize each variable and randomly sample $B$ input windows of length $L$. Given the $b$-th input window $\mathbf{X}_{b}\in\mathbb{R}^{L\times C}$, we apply instance-level normalization along the temporal dimension:
\begin{equation}
    \widetilde{\mathbf{X}}_{b}
    =
    \frac{\mathbf{X}_{b}-\operatorname{Mean}_{L}(\mathbf{X}_{b})}
    {\operatorname{Std}_{L}(\mathbf{X}_{b})+\epsilon},
\end{equation}
where $\operatorname{Mean}_{L}(\cdot)$ and $\operatorname{Std}_{L}(\cdot)$ are computed along the temporal dimension for each variable.

Then, for each variable $c$, the normalized sequence is divided into $N$ non-overlapping patches with patch length $P$:
\begin{equation}
    \mathbf{X}_{b,c}^{p}
    =
    \operatorname{Unfold}(\widetilde{\mathbf{X}}_{b}^{(c)})
    \in \mathbb{R}^{N\times P},
\end{equation}
where $N=L/P$ for non-overlapping patches, and we assume $L$ is divisible by $P$.

Each patch vector is further normalized by its $\ell_2$-norm:
\begin{equation}
    \overline{\mathbf{X}}_{b,c}^{p}[i]
    =
    \frac{\mathbf{X}_{b,c}^{p}[i]}
    {\|\mathbf{X}_{b,c}^{p}[i]\|_2+\epsilon},
    \quad i=1,\dots,N.
\end{equation}
We then construct the patch-level temporal relationship matrix as
\begin{equation}
    \mathbf{G}_{b,c}
    =
    \overline{\mathbf{X}}_{b,c}^{p}
    \left(\overline{\mathbf{X}}_{b,c}^{p}\right)^{\top}
    \in \mathbb{R}^{N\times N},
\end{equation}
where $\mathbf{G}_{b,c}[i,j]$ measures the normalized relationship strength between the $i$-th and $j$-th relative patch positions.

\paragraph{Shared temporal structures across samples and variables.}
To examine shared temporal structures, we stack the vectorized relationship matrices across samples and across variables. For the sample-side analysis, each variable $c$ is fixed and the matrices from all $B$ windows are collected; for the variable-side analysis, each sample $b$ is fixed and the matrices from all $C$ variables are collected:
\begin{equation}
\begin{aligned}
    \mathcal{G}_{c}^{(\mathrm{sample})}
    &=\left[
    \operatorname{vec}(\mathbf{G}_{1,c}),\ldots,
    \operatorname{vec}(\mathbf{G}_{B,c})
    \right]^\top
    \in\mathbb{R}^{B\times N^2},\\
    \mathcal{G}_{b}^{(\mathrm{var})}
    &=\left[
    \operatorname{vec}(\mathbf{G}_{b,1}),\ldots,
    \operatorname{vec}(\mathbf{G}_{b,C})
    \right]^\top
    \in\mathbb{R}^{C\times N^2}.
\end{aligned}
\end{equation}
Before performing SVD, the rows of both matrices are normalized to compare temporal relationship patterns on a common scale:
\begin{equation}
    \overline{\mathcal{G}}[j]
    =\frac{\mathcal{G}[j]}{\|\mathcal{G}[j]\|_2+\epsilon},
\end{equation}
where $\mathcal{G}$ denotes either stacked matrix and $j$ indexes its rows. This normalization removes differences in vector magnitude while preserving the directions of the temporal relationship representations.

Let $\{\sigma_{c,i}^{(\mathrm{sample})}\}_i$ and $\{\sigma_{b,i}^{(\mathrm{var})}\}_i$ denote the ordered singular values of $\overline{\mathcal{G}}_{c}^{(\mathrm{sample})}$ and $\overline{\mathcal{G}}_{b}^{(\mathrm{var})}$, respectively. The averaged spectral energies are defined as
\begin{equation}
    \bar{e}_{i}^{(\mathrm{sample})}
    =\frac{1}{C}\sum_{c=1}^{C}
    \left(\sigma_{c,i}^{(\mathrm{sample})}\right)^2,
    \qquad
    \bar{e}_{i}^{(\mathrm{var})}
    =\frac{1}{B}\sum_{b=1}^{B}
    \left(\sigma_{b,i}^{(\mathrm{var})}\right)^2.
\end{equation}
These spectra indicate the extent to which temporal relationship patterns are concentrated in a small number of directions across samples and variables, respectively.

\paragraph{Spectral concentration metrics.}
We quantify spectral concentration by measuring how many singular components are required to explain 95\% of the spectral energy. Given an averaged spectral energy sequence $\{\bar{e}_i\}_{i=1}^{R}$, the 95\% energy rank is defined as
\begin{equation}
    r_{95}
    =
    \min
    \left\{
    r:
    \frac{\sum_{i=1}^{r}\bar{e}_{i}}
    {\sum_{i=1}^{R}\bar{e}_{i}}
    \geq 0.95
    \right\}.
\end{equation}
Here $R$ denotes the reference dimension of the full vectorized representation:
\begin{equation}
    R=
    \begin{cases}
        \min(C,N^2), & \text{for } \mathcal{G}_{b}^{(\mathrm{var})},\\
        \min(B,N^2), & \text{for } \mathcal{G}_{c}^{(\mathrm{sample})}.
    \end{cases}
\end{equation}
To make the degree of spectral concentration more intuitive, we report the rank compression factor:
\begin{equation}
    \kappa_{95}
    =
    \frac{R}{r_{95}}.
\end{equation}
A larger $\kappa_{95}$ indicates that 95\% of the spectral energy is captured by a smaller fraction of the reference dimension, corresponding to stronger spectral concentration. However, spectral concentration can also arise from the Gram construction itself: symmetry introduces redundant coordinates in the vectorized representation, while patch normalization produces a common unit diagonal across matrices. A reference level is therefore needed to assess the additional concentration associated with temporal structure. To this end, we evaluate two null controls through the same analysis pipeline and compare their $\kappa_{95}$ values with those of the real data.

\paragraph{Quantitative analysis.}
\begin{table}[htbp]
  \caption{Shuffle denotes independent patch permutations and IID denotes independent Gaussian windows. Results are averaged over repeated runs of each null control.}
  \label{tab:spectral_concentration}
  \centering
  \small
  \setlength{\tabcolsep}{4pt}
  \renewcommand{\arraystretch}{1.12}
  \definecolor{oddrowgray}{HTML}{E7E6E6}
  \resizebox{\linewidth}{!}{%
  \begin{tabular}{l*{8}{c}}
    \toprule
    \multicolumn{9}{c}{\textbf{(a) Sample side: across temporal windows}}\\
    \midrule
    \textbf{Dataset} & \shortstack{\textbf{Patch size} $P$} & \shortstack{\textbf{Patch number} $N$} & \textbf{Variables $C$} & \textbf{Real} & \textbf{Shuffle} & \textbf{IID}
    & \textbf{Real/Shuffle} & \textbf{Real/IID}\\
    \midrule
    \rowcolor{oddrowgray}
    ETTh1       & 24 & 30 & 7   & 5.59  & 2.80 & 2.61 & $2.00\times$  & $2.14\times$\\
    ETTh2       & 24 & 30 & 7   & 5.73  & 2.68 & 2.61 & $2.14\times$  & $2.20\times$\\
    \rowcolor{oddrowgray}
    ETTm1       & 24 & 30 & 7   & 4.74  & 2.41 & 2.61 & $1.96\times$  & $1.82\times$\\
    ETTm2       & 24 & 30 & 7   & 4.27  & 2.43 & 2.61 & $1.76\times$  & $1.64\times$\\
    \rowcolor{oddrowgray}
    Weather     & 24 & 30 & 21  & 5.92  & 2.44 & 2.61 & $2.43\times$  & $2.27\times$\\
    Electricity & 24 & 30 & 321 & 90.00 & 8.04 & 2.61 & $11.20\times$ & $34.50\times$\\
    \rowcolor{oddrowgray}
    Traffic     & 24 & 30 & 862 & 64.29 & 6.12 & 2.61 & $10.50\times$ & $24.64\times$\\
    \midrule
    \multicolumn{9}{c}{\textbf{(b) Variable side: across variables within a window}}\\
    \midrule
    \textbf{Dataset} & \shortstack{\textbf{Patch size} $P$} & \shortstack{\textbf{Patch number} $N$} & \textbf{Variables $C$} & \textbf{Real} & \textbf{Shuffle} & \textbf{IID}
    & \textbf{Real/Shuffle} & \textbf{Real/IID}\\
    \midrule
    \rowcolor{oddrowgray}
    ETTh1       & 24 & 30 & 7   & 1.75   & 1.17  & 1.00 & $1.50\times$  & $1.75\times$\\
    ETTh2       & 24 & 30 & 7   & 1.40   & 1.00  & 1.00 & $1.40\times$  & $1.40\times$\\
    \rowcolor{oddrowgray}
    ETTm1       & 24 & 30 & 7   & 1.40   & 1.00  & 1.00 & $1.40\times$  & $1.40\times$\\
    ETTm2       & 24 & 30 & 7   & 1.40   & 1.00  & 1.00 & $1.40\times$  & $1.40\times$\\
    \rowcolor{oddrowgray}
    Weather     & 24 & 30 & 21  & 2.33   & 1.11  & 1.11 & $2.11\times$  & $2.11\times$\\
    Electricity & 24 & 30 & 321 & 64.20  & 12.84 & 1.66 & $5.00\times$  & $38.60\times$\\
    \rowcolor{oddrowgray}
    Traffic     & 24 & 30 & 862 & 287.33 & 8.62  & 2.90 & $33.33\times$ & $99.00\times$\\
    \bottomrule
  \end{tabular}
  }
\end{table}

Table~\ref{tab:spectral_concentration} reports the spectral concentration statistics under $B=2000$, $L=720$, and $P=24$ ($N=30$), together with two null controls evaluated through the same pipeline. \textbf{(1) Patch shuffle} evaluates the role of temporal ordering by independently permuting patch positions for each sample-variable pair, giving $\mathbf{G}_{b,c}^{\mathrm{shuffle}}=\mathbf{\Pi}_{b,c}\mathbf{G}_{b,c}\mathbf{\Pi}_{b,c}^{\top}$. This preserves the spectrum and similarity values of each Gram matrix while disrupting the correspondence between relative patch positions across matrices. Comparing the resulting concentration with that of real data therefore assesses whether the observed concentration depends on temporal relationships recurring at aligned patch positions. \textbf{(2) IID Gaussian} provides a complementary reference by replacing the observed windows with independent Gaussian noise. Applying the same analysis to these inputs gives the spectral concentration that can arise without temporal dependencies, against which the additional concentration in real data can be assessed.

Table~\ref{tab:spectral_concentration} reveals four findings:

\noindent\textbf{(i) Shared structures at aligned temporal positions.} Across all seven datasets, Real/Shuffle is substantially greater than one on both the sample and variable sides. The reduction in concentration after shuffling supports shared temporal relationship patterns that recur at corresponding relative patch positions across windows and variables.

\noindent\textbf{(ii) Concentration beyond the random-input reference.} Real/IID is also substantially greater than one across all datasets on both sides. Temporal relationships in real time series are therefore more concentrated than those obtained from independent random inputs under the same analysis pipeline, supporting their representation through a compact set of shared patterns.

\noindent\textbf{(iii) Larger gains in Electricity and Traffic.} Electricity and Traffic exhibit substantially larger Real/Shuffle and Real/IID gains than ETT and Weather on both sides. Their temporal relationships show more pronounced concentration relative to their respective controls, with the larger Real/Shuffle gains indicating a greater dependence of this concentration on the original patch-position alignment.

\noindent\textbf{(iv) Shared regularities preserved under shuffling.} For Electricity and Traffic, shuffled-data concentration remains well above the corresponding IID reference on both sides, whereas the two controls yield closer values for ETT and Weather. This suggests that shared regularities preserved under patch permutation coexist with reusable temporal relationships tied to the original relative positions.

\paragraph{From temporal relationships to dependency parameterization.}
The above analysis provides evidence of shared temporal relationship structures across windows and variables. To connect these historical relationships with forecasting, we consider an idealized model in which historical and future patch representations are related by a shared temporal evolution operator.

For a sample-variable context $i$, let $\mathbf{X}_i\in\mathbb{R}^{N\times d}$ and $\mathbf{Y}_i\in\mathbb{R}^{M\times d}$ denote row-normalized historical and future patch representations in a common state space, where $N=L/P$ and $M=H/P$, assuming both lengths are divisible by $P$. Suppose
\begin{equation}
    \mathbf{Y}_i=\mathbf{T}\mathbf{X}_i+\boldsymbol{\varepsilon}_i,
    \qquad
    \mathbb{E}[\boldsymbol{\varepsilon}_i\mid\mathbf{X}_i]=0,
    \qquad
    \mathbf{T}\in\mathbb{R}^{M\times N},
\end{equation}
where $\mathbf{T}$ is shared across the contexts under consideration and $\boldsymbol{\varepsilon}_i$ captures residual variation. The historical auto-relation and history--future cross-relation then satisfy
\begin{equation}
    \mathbf{G}_i=\mathbf{X}_i\mathbf{X}_i^\top,
    \qquad
    \boldsymbol{\Gamma}_i=\mathbf{Y}_i\mathbf{X}_i^\top
    =\mathbf{T}\mathbf{G}_i+\boldsymbol{\varepsilon}_i\mathbf{X}_i^\top,
\end{equation}
and hence
$\mathbb{E}[\boldsymbol{\Gamma}_i\mid\mathbf{X}_i]
=\mathbf{T}\mathbf{G}_i$.

Motivated by the observed spectral concentration, suppose the historical relation family admits an approximate representation through a mean structure and $r$ shared directions:
\begin{equation}
    \mathbf{G}_i=\boldsymbol{\mu}_G
    +\sum_{\ell=1}^{r}\beta_{i,\ell}\mathbf{Q}_{\ell}
    +\boldsymbol{\Delta}_i,
\end{equation}
where $\boldsymbol{\Delta}_i$ denotes the approximation residual. It follows that
\begin{equation}
    \boldsymbol{\Gamma}_i
    =\mathbf{T}\boldsymbol{\mu}_G
    +\sum_{\ell=1}^{r}\beta_{i,\ell}(\mathbf{T}\mathbf{Q}_{\ell})
    +\left(\mathbf{T}\boldsymbol{\Delta}_i
    +\boldsymbol{\varepsilon}_i\mathbf{X}_i^\top\right).
\end{equation}
Thus, under shared temporal evolution, the historical directions $\mathbf{Q}_{\ell}$ give rise to shared directions $\mathbf{T}\mathbf{Q}_{\ell}$ in the history--future relation family. When the residual terms are small, the same coefficients $\beta_{i,\ell}$ organize both families. This provides a conditional link between reusable historical relationships and a compact description of relationships relevant to forecasting.

Motivated by this connection, TStruct parameterizes temporal aggregation through shared structure matrices:
\begin{equation}
    \mathbf{A}_i
    =\operatorname{Softmax}_{\mathrm{row}}\left(
        \sum_{k=1}^{K}\alpha_{i,k}\mathbf{B}_k\right),
    \qquad
    \widetilde{\mathbf{Z}}_i=\mathbf{A}_i\mathbf{Z}_i,
\end{equation}
where $\mathbf{Z}_i$ denotes projected historical features. Learned jointly across training observations under forecasting supervision, the shared matrices $\mathbf{B}_k$ define a compact family of temporal aggregation operators $\mathbf{A}_i$ on historical patch coordinates. This parameterization is intended to capture stable temporal dependency patterns relevant to forecasting, while the forecasting head maps the aggregated features to the future horizon. The connection motivates the shared structural parameterization of $\mathbf{A}_i$ without requiring equality with $\mathbf{G}_i$ or $\boldsymbol{\Gamma}_i$, or inheritance of their spectra. By drawing on patterns shared across observations, TStruct aims to reduce reliance on incidental relationships in noisy time series and improve forecasting generalization. These intended benefits are evaluated in Section~\ref{sec:genera} and Appendix~\ref{appendix:noisy}.

\section{Implementation Details}
\label{appendix:implementation_details}

All experiments performed in this study are conducted on a workstation equipped with NVIDIA RTX PRO 6000 GPUs with 96GB memory. TStruct is implemented in PyTorch 2.8.0. We fix the random seed to $2026$ for our runs. Following the standard evaluation protocol~\citep{tfb}, we set \textit{drop\_last=False} for the data loader during testing to ensure that all test samples are evaluated.

For the main forecasting experiments, TStruct uses a fixed input look-back length of $L=720$. We train TStruct using the Adam optimizer~\citep{adam} for at most 25 epochs with an early stopping patience of 10 epochs. The best checkpoint is selected according to the validation loss.

\paragraph{Baseline configurations.}
To compare the best performance of each method within the evaluated configurations, we retain the training objectives used by the respective methods and allow look-back selection where specified below. The sources and configurations of the results in Table~\ref{tab:summary_results} are as follows: \textbf{TStruct.} We report our results using the Huber+FFT objective described below and a fixed look-back length of $L=720$. \textbf{DUET.} We use the paper reported results obtained with the L1 objective, selecting the best performing look-back length from $L\in\{96,192,336,512,720\}$ for each dataset horizon setting. \textbf{PhaseFormer.} We use the paper reported results obtained with the Huber objective and $L=720$. \textbf{Other baselines.} We run their released implementations with their native training objectives and select the best performing look-back length from $L\in\{96,192,336,512,720\}$ for each dataset horizon setting.

\paragraph{Training objective.}
We optimize TStruct with a hybrid objective that combines Huber loss and FFT loss~\citep{fredf}. Given the prediction $\hat{\mathbf{Y}}$ and ground truth $\mathbf{Y}$, the Huber loss is defined as
\begin{equation}
    \mathcal{L}_{\mathrm{Huber}}
    =
    \frac{1}{|\Omega|}
    \sum_{i\in\Omega}
    \ell_{\delta}(\hat{y}_{i}-y_i),
\end{equation}
where $\Omega$ denotes all predicted entries and we set $\delta=0.1$ in all experiments. The element-wise Huber loss is defined as
\begin{equation}
    \ell_{\delta}(e)
    =
    \begin{cases}
    \frac{1}{2}e^2, & |e|\leq \delta,\\
    \delta\left(|e|-\frac{1}{2}\delta\right), & |e|>\delta.
    \end{cases}
\end{equation}
The FFT loss measures the discrepancy between predictions and ground truth in the frequency domain:
\begin{equation}
    \mathcal{L}_{\mathrm{FFT}}
    =
    \left\|
    \operatorname{rFFT}(\hat{\mathbf{Y}}, \mathrm{dim}=1)
    -
    \operatorname{rFFT}(\mathbf{Y}, \mathrm{dim}=1)
    \right\|_{1},
\end{equation}
where $\operatorname{rFFT}(\cdot,\mathrm{dim}=1)$ denotes the real-valued fast Fourier transform along the temporal dimension, and $\|\cdot\|_1$ is implemented as the mean absolute value over all frequency-domain entries. The final training objective is
\begin{equation}
    \mathcal{L}
    =
    \mathcal{L}_{\mathrm{Huber}}
    +
    \lambda_{\mathrm{FFT}}\mathcal{L}_{\mathrm{FFT}},
\end{equation}
where $\lambda_{\mathrm{FFT}}$ is selected from $\{0.0,0.1,0.2,0.3\}$.

\paragraph{Evaluation metrics.}
Following prior LTSF studies~\citep{informer,autoformer}, we use Mean Squared Error (MSE) and Mean Absolute Error (MAE) as evaluation metrics:
\begin{equation}
    \mathrm{MSE}
    =
    \frac{1}{|\Omega|}
    \sum_{i\in\Omega}
    (\hat{y}_i-y_i)^2,
    \quad
    \mathrm{MAE}
    =
    \frac{1}{|\Omega|}
    \sum_{i\in\Omega}
    |\hat{y}_i-y_i|.
\end{equation}

\paragraph{Hyperparameter search space.}
We tune the hyperparameters over the following search space. The patch size is selected from
$P \in \{24,48,72,120,180,360\}$, the number of encoder layers from
$\{1,3\}$, the hidden dimension from
$D \in \{128,256,512\}$, and the number of shared temporal structures from
$K \in \{1,2,4,5,8,16,32\}$.
For regularization, the MLP dropout rate is selected from
$\{0.1,0.2,0.3,0.4\}$, the TSM dropout rate from
$\{0.0,0.1,0.2,0.3,0.4\}$, and the PCAhead dropout rate from
$\{0.0,0.1,0.2\}$.
For optimization, the batch size is selected from
$\{16,64,256,512\}$, and the learning rate $\eta$ is selected from
$\{0.001,0.0015,0.002,0.005,0.015\}$.

\section{Additional Experimental Analyses}
\label{appendix:robust_sensitivity}

\subsection{Random Seed Stability}
\label{appendix:seed}
\begin{table*}[htbp]
  \caption{Stability analysis under different random seeds.}
  \label{tab:seed_stability}
  \centering
  \scriptsize
  \setlength{\tabcolsep}{3pt}
  \renewcommand{\arraystretch}{1.05}
  \definecolor{headerblue}{HTML}{D9E1EC}
  \definecolor{metricgray}{HTML}{E7E6E6}

  \resizebox{\textwidth}{!}{
  \begin{tabular}{c|c|cc|cc|cc|cc|cc|cc|cc}
  \toprule[1.1pt]
  \rowcolor{headerblue}
  \multicolumn{2}{c|}{\textbf{Seed}} 
  & \multicolumn{2}{c|}{\textbf{2024}}
  & \multicolumn{2}{c|}{\textbf{2025}}
  & \multicolumn{2}{c|}{\textbf{2026}}
  & \multicolumn{2}{c|}{\textbf{2027}}
  & \multicolumn{2}{c|}{\textbf{2028}}
  & \multicolumn{2}{c|}{\textbf{Mean}}
  & \multicolumn{2}{c}{\textbf{Std}} \\
  \cmidrule(lr){1-2}
  \cmidrule(lr){3-4}
  \cmidrule(lr){5-6}
  \cmidrule(lr){7-8}
  \cmidrule(lr){9-10}
  \cmidrule(lr){11-12}
  \cmidrule(lr){13-14}
  \cmidrule(lr){15-16}
  \rowcolor{metricgray}
  \textbf{Dataset} & \textbf{Metric}
  & \textbf{MSE} & \textbf{MAE}
  & \textbf{MSE} & \textbf{MAE}
  & \textbf{MSE} & \textbf{MAE}
  & \textbf{MSE} & \textbf{MAE}
  & \textbf{MSE} & \textbf{MAE}
  & \textbf{MSE} & \textbf{MAE}
  & \textbf{MSE} & \textbf{MAE} \\
  \midrule

    \multirow{5}{*}{ETTh1}
    & 96  & 0.345 & 0.380 & 0.344 & 0.380 & 0.345 & 0.381 & 0.344 & 0.390 & 0.345 & 0.380 & 0.345 & 0.382 & 0.001 & 0.004 \\
    & 192 & 0.382 & 0.404 & 0.383 & 0.406 & 0.383 & 0.405 & 0.384 & 0.406 & 0.384 & 0.406 & 0.383 & 0.405 & 0.001 & 0.001 \\
    & 336 & 0.407 & 0.421 & 0.405 & 0.421 & 0.405 & 0.420 & 0.407 & 0.422 & 0.407 & 0.421 & 0.406 & 0.421 & 0.001 & 0.001 \\
    & 720 & 0.424 & 0.453 & 0.426 & 0.453 & 0.431 & 0.456 & 0.428 & 0.456 & 0.424 & 0.450 & 0.427 & 0.454 & 0.003 & 0.003 \\
    \cmidrule(lr){2-16}
    & \emph{Avg.}
          & \textbf{0.390} & \textbf{0.415}
          & \textbf{0.390} & \textbf{0.415}
          & \textbf{0.391} & \textbf{0.416}
          & \textbf{0.391} & \textbf{0.419}
          & \textbf{0.390} & \textbf{0.414}
          & \textbf{0.390} & \textbf{0.416}
          & \textbf{0.001} & \textbf{0.002} \\
    \midrule

    \multirow{5}{*}{Weather}
    & 96  & 0.137 & 0.179 & 0.137 & 0.179 & 0.137 & 0.181 & 0.138 & 0.180 & 0.138 & 0.180 & 0.137 & 0.180 & 0.001 & 0.001 \\
    & 192 & 0.181 & 0.225 & 0.182 & 0.225 & 0.181 & 0.226 & 0.180 & 0.225 & 0.181 & 0.225 & 0.181 & 0.225 & 0.001 & 0.000 \\
    & 336 & 0.231 & 0.267 & 0.231 & 0.268 & 0.230 & 0.267 & 0.231 & 0.269 & 0.230 & 0.268 & 0.231 & 0.268 & 0.001 & 0.001 \\
    & 720 & 0.306 & 0.325 & 0.305 & 0.325 & 0.304 & 0.324 & 0.305 & 0.325 & 0.305 & 0.326 & 0.305 & 0.325 & 0.001 & 0.001 \\
    \cmidrule(lr){2-16}
    & \emph{Avg.}
          & \textbf{0.214} & \textbf{0.249}
          & \textbf{0.214} & \textbf{0.249}
          & \textbf{0.213} & \textbf{0.250}
          & \textbf{0.214} & \textbf{0.250}
          & \textbf{0.214} & \textbf{0.250}
          & \textbf{0.214} & \textbf{0.249}
          & \textbf{0.000} & \textbf{0.000} \\
    \bottomrule[1.1pt]
  \end{tabular}
  }
\end{table*}

To examine the stability of TStruct under different random initializations and stochastic training processes, we repeat experiments with five random seeds, i.e., $2024, 2025, 2026, 2027,$ and $2028$. Table~\ref{tab:seed_stability} reports the results on ETTh1 and Weather, including the mean and standard deviation across seeds.

The results show that TStruct is highly stable across different random seeds. The standard deviation is consistently very small, mostly ranging from $0.000$ to $0.002$ for both MSE and MAE. This indicates that the performance of TStruct is not sensitive to random initialization or stochastic optimization, and the observed improvements are reproducible rather than caused by favorable random seeds.

\subsection{Initialization Study}
\label{appendix:Initialization}
\begin{table*}[htbp]
  \caption{Sensitivity analysis under different initialization strategies.}
  \label{tab:initialization_sensitivity}
  \centering
  \small
  \setlength{\tabcolsep}{3pt}
  \renewcommand{\arraystretch}{0.95}
  \definecolor{headerblue}{HTML}{D9E1EC}
  \definecolor{metricgray}{HTML}{E7E6E6}

  \begin{tabular}{c|c|cc|cc|cc|cc}
  \toprule[1.1pt]
  \rowcolor{headerblue}
  \multicolumn{2}{c|}{\textbf{Initialization}}
  & \multicolumn{2}{c|}{\textbf{Random}}
  & \multicolumn{2}{c|}{\textbf{Orthogonal}}
  & \multicolumn{2}{c|}{\textbf{Xavier}}
  & \multicolumn{2}{c}{\textbf{Kaiming}} \\
  \cmidrule(lr){1-2}
  \cmidrule(lr){3-4}
  \cmidrule(lr){5-6}
  \cmidrule(lr){7-8}
  \cmidrule(lr){9-10}
  \rowcolor{metricgray}
  \textbf{Dataset} & \textbf{Metric}
  & \textbf{MSE} & \textbf{MAE}
  & \textbf{MSE} & \textbf{MAE}
  & \textbf{MSE} & \textbf{MAE}
  & \textbf{MSE} & \textbf{MAE} \\
  \midrule

    \multirow{5}{*}{ETTh1}
    & 96  & 0.366 & 0.398 & 0.345 & 0.380 & 0.345 & 0.380 & 0.345 & 0.380 \\
    & 192 & 0.421 & 0.439 & 0.383 & 0.405 & 0.383 & 0.405 & 0.383 & 0.405 \\
    & 336 & 0.426 & 0.442 & 0.407 & 0.420 & 0.407 & 0.420 & 0.407 & 0.420 \\
    & 720 & 0.497 & 0.503 & 0.433 & 0.456 & 0.434 & 0.459 & 0.430 & 0.458 \\
    \cmidrule(lr){2-10}
    & \emph{Avg.}
          & \textbf{0.428} & \textbf{0.446}
          & \textbf{0.392} & \textbf{0.415}
          & \textbf{0.392} & \textbf{0.416}
          & \textbf{0.391} & \textbf{0.416} \\
    \midrule

    \multirow{5}{*}{Weather}
    & 96  & 0.142 & 0.187 & 0.137 & 0.181 & 0.137 & 0.180 & 0.137 & 0.180 \\
    & 192 & 0.187 & 0.230 & 0.181 & 0.226 & 0.182 & 0.226 & 0.181 & 0.225 \\
    & 336 & 0.247 & 0.282 & 0.230 & 0.267 & 0.231 & 0.267 & 0.230 & 0.268 \\
    & 720 & 0.327 & 0.347 & 0.304 & 0.324 & 0.304 & 0.325 & 0.305 & 0.325 \\
    \cmidrule(lr){2-10}
    & \emph{Avg.}
          & \textbf{0.226} & \textbf{0.262}
          & \textbf{0.213} & \textbf{0.250}
          & \textbf{0.214} & \textbf{0.250}
          & \textbf{0.213} & \textbf{0.250} \\
    \bottomrule[1.1pt]
  \end{tabular}
\end{table*}
We study the effect of different initialization strategies for the shared temporal structures $\mathcal{B}$. Specifically, we compare Random, Orthogonal, Xavier, and Kaiming initialization, as reported in Table~\ref{tab:initialization_sensitivity}. The results show that the Random baseline yields noticeably higher forecasting errors on both ETTh1 and Weather, whereas Orthogonal, Xavier, and Kaiming achieve consistently strong and comparable performance.

All initialization strategies are applied directly to the $K\times N\times N$ basis tensor. Orthogonal initialization gives the vectorized bases mutually orthogonal directions with unit Frobenius norm, while Xavier and Kaiming scale the sampling variance according to the tensor dimensions. These strategies control the initial magnitude of the dependency scores, which are subsequently converted into aggregation weights through softmax. This provides a plausible explanation for their favorable performance: appropriately scaled initial scores moderate arbitrary preferences over patch positions, allowing temporal dependency patterns to develop through forecasting supervision. The role of initialization scale is also consistent with observations on learning dynamics in deep linear networks~\citep{rhee-2026-deep-linear-nets}.

We use Orthogonal initialization as the default. It provides diverse initial structure directions with a common scale, reducing initial redundancy among the shared bases while achieving consistently strong empirical performance.

\subsection{Training Objective Comparison}
\label{appendix:training_objective}
We investigate the effect of the training objective on forecasting performance by evaluating TStruct, DUET, and PhaseFormer under the same two objectives, Huber+FFT and plain MSE. The remaining experimental settings follow
Appendix~\ref{appendix:implementation_details}.

\begin{table*}[htbp]
  \caption{Forecasting performance under two training objectives with $L=720$ for all methods. The lowest MSE and MAE for each dataset within each objective are highlighted in bold.}
  \label{tab:training_objective_comparison}
  \centering
  \small
  \setlength{\tabcolsep}{4pt}
  \renewcommand{\arraystretch}{1.12}
  \definecolor{lossblue}{HTML}{C5D3E4}
  \definecolor{headerblue}{HTML}{D9E1EC}
  \definecolor{metricgray}{HTML}{E7E6E6}
  \resizebox{\linewidth}{!}{%
  \begin{tabular}{l|cc|cc|cc||cc|cc|cc}
    \toprule[1.1pt]
    \rowcolor{lossblue}
    \textbf{Loss} & \multicolumn{6}{c||}{\textbf{Huber+FFT}} & \multicolumn{6}{c}{\textbf{Plain MSE}}\\
    \cmidrule(lr){1-1}\cmidrule(lr){2-7}\cmidrule(lr){8-13}
    \rowcolor{headerblue}
    \textbf{Method} & \multicolumn{2}{c|}{\textbf{TStruct}} & \multicolumn{2}{c|}{\textbf{DUET}} & \multicolumn{2}{c||}{\textbf{PhaseFormer}} & \multicolumn{2}{c|}{\textbf{TStruct}} & \multicolumn{2}{c|}{\textbf{DUET}} & \multicolumn{2}{c}{\textbf{PhaseFormer}}\\
    \cmidrule(lr){1-1}\cmidrule(lr){2-3}\cmidrule(lr){4-5}\cmidrule(lr){6-7}\cmidrule(lr){8-9}\cmidrule(lr){10-11}\cmidrule(lr){12-13}
    \rowcolor{metricgray}
    \textbf{Datasets} & \textbf{MSE} & \textbf{MAE} & \textbf{MSE} & \textbf{MAE} & \textbf{MSE} & \textbf{MAE} & \textbf{MSE} & \textbf{MAE} & \textbf{MSE} & \textbf{MAE} & \textbf{MSE} & \textbf{MAE}\\
    \midrule
    ETTh1 & \textbf{0.390} & \textbf{0.415} & 0.430 & 0.450 & 0.417 & 0.425 & \textbf{0.410} & \textbf{0.432} & 0.425 & 0.447 & 0.431 & 0.437\\
    \midrule[0.3pt]
    ETTh2 & \textbf{0.331} & \textbf{0.381} & 0.364 & 0.410 & 0.374 & 0.401 & \textbf{0.349} & \textbf{0.400} & 0.358 & 0.405 & 0.358 & 0.402\\
    \midrule[0.3pt]
    ETTm1 & \textbf{0.337} & \textbf{0.367} & 0.432 & 0.426 & 0.351 & 0.378 & \textbf{0.345} & \textbf{0.373} & 0.382 & 0.409 & 0.358 & 0.384\\
    \midrule[0.3pt]
    ETTm2 & \textbf{0.243} & \textbf{0.304} & 0.282 & 0.342 & 0.253 & 0.314 & \textbf{0.246} & \textbf{0.311} & 0.279 & 0.338 & 0.266 & 0.329\\
    \midrule[0.3pt]
    Weather & \textbf{0.213} & \textbf{0.249} & 0.218 & 0.252 & 0.231 & 0.271 & \textbf{0.217} & \textbf{0.258} & 0.249 & 0.293 & 0.231 & 0.275\\
    \midrule[0.3pt]
    Electricity & \textbf{0.151} & \textbf{0.244} & 0.158 & 0.251 & 0.159 & 0.249 & \textbf{0.157} & \textbf{0.251} & 0.158 & 0.256 & 0.161 & 0.252\\
    \midrule[0.3pt]
    Traffic & \textbf{0.381} & 0.252 & 0.387 & 0.267 & 0.387 & \textbf{0.249} & \textbf{0.385} & 0.264 & 0.388 & 0.270 & 0.393 & \textbf{0.259}\\
    \bottomrule[1.1pt]
  \end{tabular}%
  }
\end{table*}

Table~\ref{tab:training_objective_comparison} shows that TStruct ranks first in MSE on all seven datasets and in MAE on six under both objectives, with PhaseFormer leading in MAE on Traffic. Huber+FFT further improves TStruct's MSE and MAE across all seven datasets compared with plain MSE. Its strong performance under plain MSE supports the effectiveness of shared temporal structure modeling, while the hybrid objective provides additional gains.

\subsection{Noise Injection Robustness}
\label{appendix:noisy}
To evaluate robustness under local abnormal perturbations, we conduct noise injection experiments on ETTh1. For each normalized input window, whose values are standardized to approximately zero mean and unit variance, we randomly select a continuous segment and inject Gaussian noise into this segment. The noise ratio controls the length of the perturbed segment relative to the input window, while the noise strength controls the magnitude of the injected Gaussian noise. The start position of the noisy segment is randomly sampled within each input window.

We compare TSM and MHA across different noise ratios and strengths, with the results summarized in Table~\ref{tab:noise_injection}. TSM achieves lower MSE and MAE throughout both experiments. As the noise ratio increases from $0.05$ to $0.30$, its MSE ranges from $0.392$ to $0.400$, remaining $0.042$--$0.047$ below MHA. At a noise strength of $8$, the MSE gap reaches $0.075$, with TSM obtaining $0.404$ compared with $0.479$ for MHA.

\begin{table*}[htbp]
  \caption{Noise injection robustness on ETTh1 under varying noise ratios (left) and strengths (right). $\Delta$ denotes the error reduction of TSM relative to MHA.}
  \label{tab:noise_injection}
  \centering
  \small
  \setlength{\tabcolsep}{4pt}
  \renewcommand{\arraystretch}{1.12}
  \definecolor{lossblue}{HTML}{C5D3E4}
  \definecolor{headerblue}{HTML}{D9E1EC}
  \definecolor{metricgray}{HTML}{E7E6E6}
  \resizebox{\linewidth}{!}{%
  \begin{tabular}{c|cc|cc|cc||c|cc|cc|cc}
    \toprule[1.1pt]
    \rowcolor{lossblue}
    \textbf{Setting} & \multicolumn{6}{c||}{\textbf{Varying noise ratio}} & \textbf{Setting} & \multicolumn{6}{c}{\textbf{Varying noise strength}}\\
    \cmidrule(lr){1-7}\cmidrule(lr){8-14}
    \rowcolor{headerblue}
    \textbf{Method} & \multicolumn{2}{c|}{\textbf{TSM}} & \multicolumn{2}{c|}{\textbf{MHA}} & \multicolumn{2}{c||}{\textbf{$\Delta$}} & \textbf{Method} & \multicolumn{2}{c|}{\textbf{TSM}} & \multicolumn{2}{c|}{\textbf{MHA}} & \multicolumn{2}{c}{\textbf{$\Delta$}}\\
    \cmidrule(lr){1-1}\cmidrule(lr){2-3}\cmidrule(lr){4-5}\cmidrule(lr){6-7}\cmidrule(lr){8-8}\cmidrule(lr){9-10}\cmidrule(lr){11-12}\cmidrule(lr){13-14}
    \rowcolor{metricgray}
    \textbf{Ratio} & \textbf{MSE} & \textbf{MAE} & \textbf{MSE} & \textbf{MAE} & \textbf{MSE} & \textbf{MAE} & \textbf{Strength} & \textbf{MSE} & \textbf{MAE} & \textbf{MSE} & \textbf{MAE} & \textbf{MSE} & \textbf{MAE}\\
    \midrule
    0.05 & 0.392 & 0.417 & 0.434 & 0.449 & 0.042 & 0.033 & 2 & 0.393 & 0.417 & 0.437 & 0.452 & 0.044 & 0.035 \\
    \midrule[0.3pt]
    0.10 & 0.394 & 0.418 & 0.441 & 0.455 & 0.047 & 0.036 & 3 & 0.394 & 0.418 & 0.441 & 0.455 & 0.047 & 0.036 \\
    \midrule[0.3pt]
    0.20 & 0.397 & 0.425 & 0.443 & 0.457 & 0.046 & 0.033 & 5 & 0.396 & 0.420 & 0.434 & 0.449 & 0.038 & 0.029 \\
    \midrule[0.3pt]
    0.30 & 0.400 & 0.427 & 0.442 & 0.458 & 0.042 & 0.031 & 8 & 0.404 & 0.426 & 0.479 & 0.473 & 0.075 & 0.047 \\
    \bottomrule[1.1pt]
  \end{tabular}%
  }
\end{table*}

These results support the robustness of shared temporal structure modeling under local perturbations. By aggregating temporal information through structures learned across observations, TSM provides a shared basis for processing corrupted inputs, which may help explain its consistently lower forecasting errors across the tested noise conditions.

\subsection{Hyperparameter Sensitivity}
\label{appendix:hyperparameter_sensitivity}
We further analyze the sensitivity of TStruct to two key hyperparameters: the number of shared temporal structures $K$ and the number of patches $N$. Experiments are conducted on the ETT benchmarks under different prediction horizons.

\begin{figure*}[p]
    \centering
    \includegraphics[width=\linewidth,height=0.42\textheight,keepaspectratio]{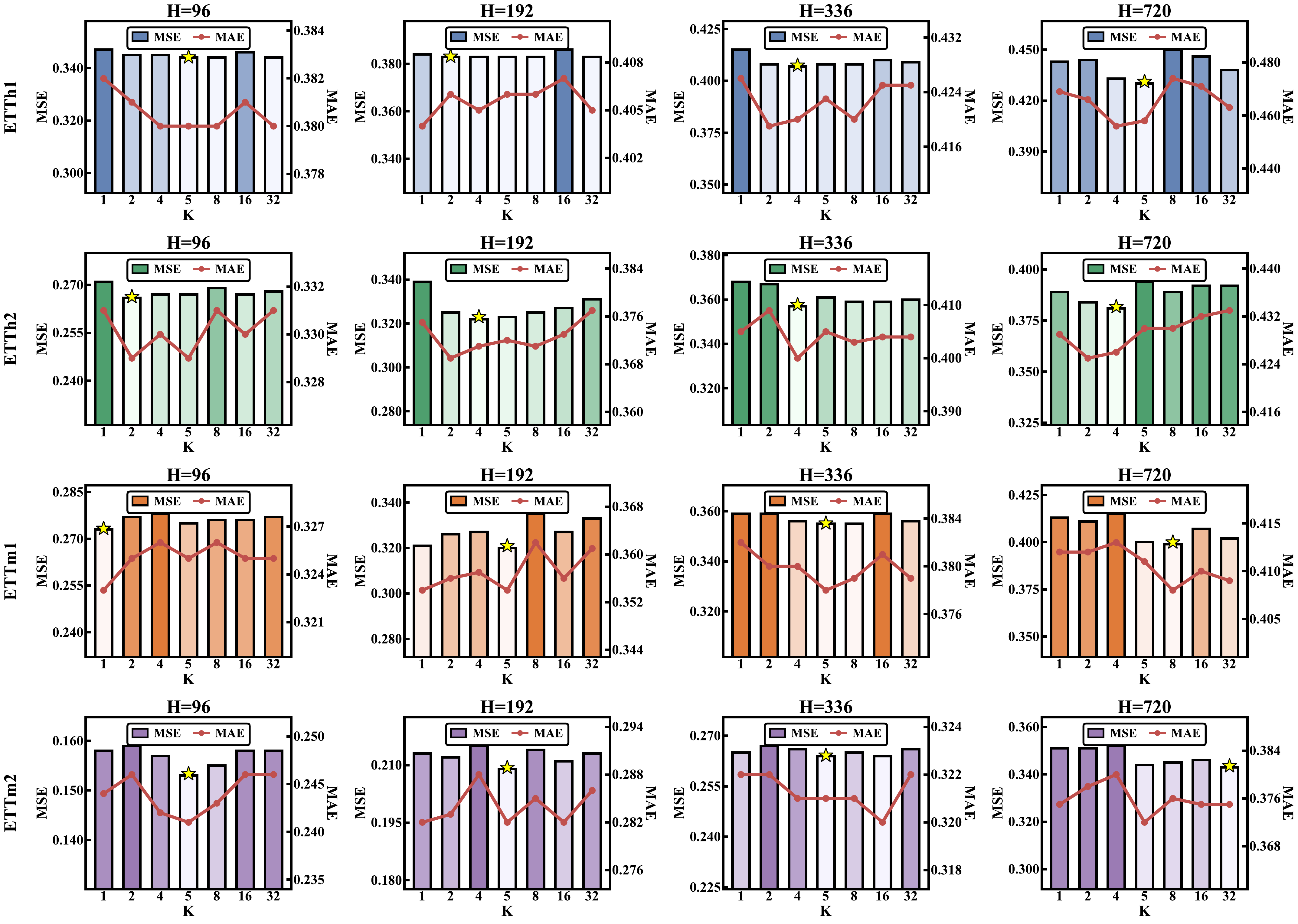}
    \caption{Hyperparameter sensitivity analysis for the number of shared temporal structures $K$ on the ETT benchmarks. Bar plots denote MSE and line plots denote MAE under different prediction horizons.}
    \label{fig:K_parameter}

    \vspace{2em}

    \includegraphics[width=\linewidth,height=0.42\textheight,keepaspectratio]{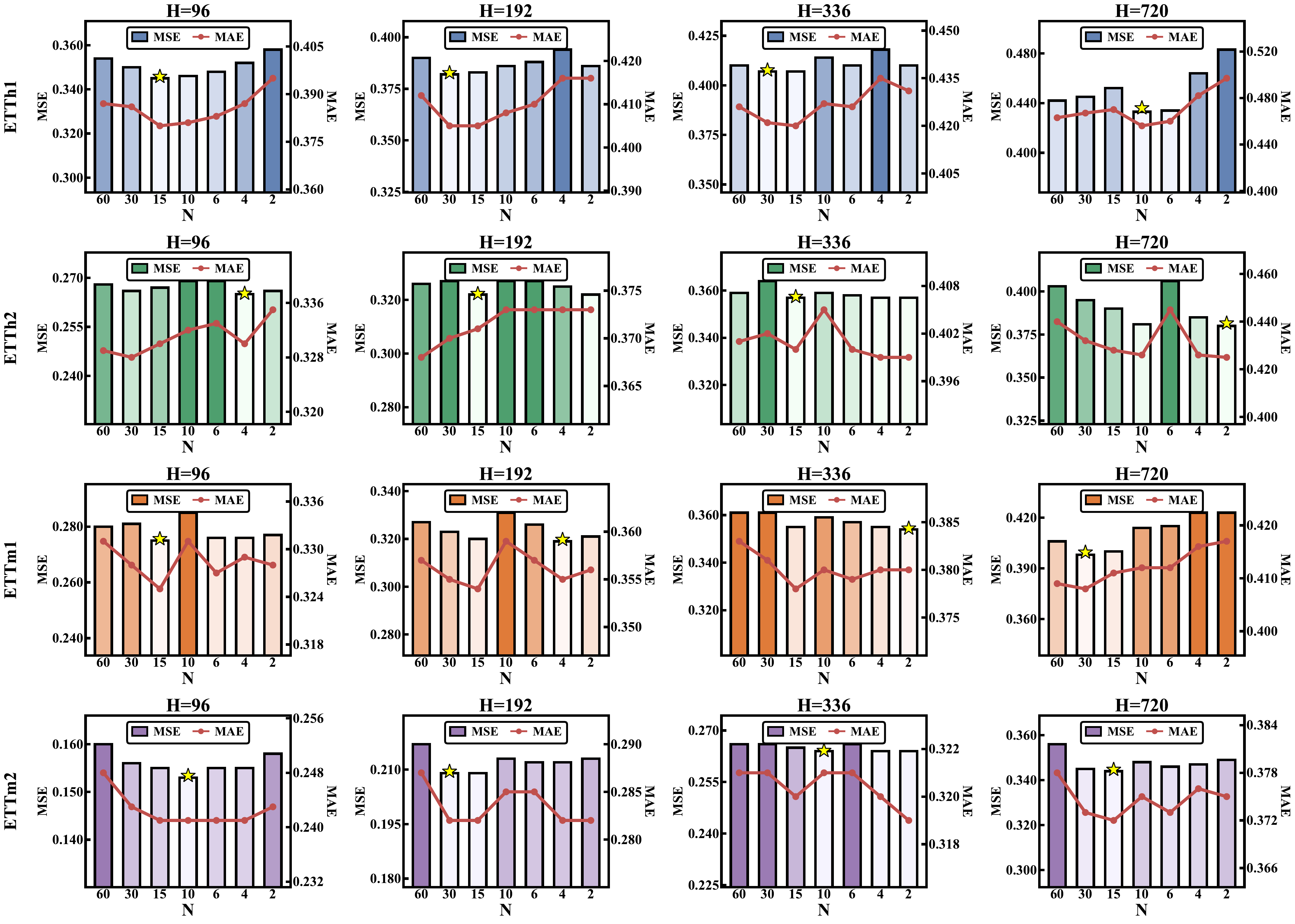}
    \caption{Hyperparameter sensitivity analysis for the number of patches $N$ on the ETT benchmarks. TStruct remains stable within a broad middle range of patch-level resolutions, while overly small or overly large $N$ may lead to underfitting or overfitting.}
    \label{fig:N_parameter}
\end{figure*}

Figure~\ref{fig:K_parameter} reports the forecasting performance under different values of $K$. Overall, TStruct remains stable across a wide range of $K$. A very small $K$ may limit the diversity of shared temporal structures, making it difficult to represent heterogeneous temporal dependencies. In contrast, an overly large $K$ may introduce redundant structures and increase optimization difficulty. The results show that a moderate number of shared structures is sufficient for strong performance, and we use $K=4$ as the default setting in our main experiments.

Figure~\ref{fig:N_parameter} shows the sensitivity to the number of patches $N$, which controls the resolution of the patch-level temporal dependency matrix. TStruct achieves robust performance within a broad middle range, especially when $N \in [10,30]$. When $N$ is too small, the temporal relationship matrix becomes overly coarse and may underfit fine-grained temporal dependencies. When $N$ is too large, each patch contains fewer time points, which may make the learned dependency structures more sensitive to local noise and increase the risk of overfitting. These results indicate that TStruct is robust to reasonable patch-level resolutions while benefiting from a balanced choice of $N$.

\subsection{Efficiency and Complexity Analysis}
\label{appendix:efficiency_complexity}
We provide an additional efficiency analysis on the Electricity dataset to evaluate the accuracy-efficiency trade-off of TStruct. Since TStruct models temporal dependencies at the patch level, we compare it with representative patch-based Transformer baselines, including Crossformer and PatchTST. All models are evaluated under the same setting with input length $L=720$, patch size and stride $P=S=48$, and hidden dimension $D=512$. We report forecasting accuracy, parameter number, multiply-accumulate operations (MACs), GPU memory usage, training time per step, and inference time per batch in Table~\ref{tab:efficiency_comparison}.

\paragraph{Complexity analysis.}
The efficiency of TStruct mainly comes from its structured temporal dependency modeling. In attention-based patch-level temporal mixing, the dependency score matrix is computed from input-dependent query-key interactions, which introduces a score estimation cost of $O(N^2D)$, where $N$ is the number of patches and $D$ is the hidden dimension. In contrast, TSM constructs temporal dependency scores by combining $K$ shared temporal structures, with a score construction cost of $O(N^2K)$. Since $K \ll D$, TSM provides a more compact way to estimate temporal dependency scores and avoids fully input-dependent pairwise attention computation.

After obtaining the normalized dependency matrix, TSM still performs patch-level feature aggregation, so the overall temporal mixing remains quadratic in $N$. The key difference is that TSM replaces unconstrained input-dependent attention score estimation with shared-adaptive structure composition. This makes temporal dependency modeling more structured and computationally efficient while preserving sample- and variable-specific adaptability.

\paragraph{Empirical efficiency.}
As shown in Table~\ref{tab:efficiency_comparison}, TStruct achieves the best MSE while requiring fewer parameters, lower MACs, less GPU memory, and faster training and inference than the patch-based Transformer baselines. Compared with PatchTST, TStruct reduces parameters from $5.496$M to $3.324$M and MACs from $23.134$G to $11.779$G, while reducing MSE from $0.171$ to $0.151$. Compared with Crossformer, TStruct also achieves substantially lower memory usage and faster runtime. These results show that TStruct improves forecasting accuracy with a favorable efficiency profile by replacing fully input-dependent attention with shared-adaptive temporal dependency modeling.
\begin{table}[H]
    \centering
    \caption{
    Efficiency comparison on the Electricity dataset.
    Crossformer~\citep{crossformer} and PatchTST~\citep{patchtst} are included as representative patch-based Transformer baselines.
    All models are evaluated under the same setting.
    We report MSE, parameters, MACs, GPU memory, training time per step, and inference time per batch.
    \textbf{Bold} indicates the best result, and \underline{underline} indicates the second-best result.
    }
    \label{tab:efficiency_comparison}
    \renewcommand{\arraystretch}{1.2}
    \resizebox{\linewidth}{!}{
    \begin{tabular}{c|c|c|c|c|c|c}
        \toprule[1.1pt]
        \textbf{Model} 
        & \textbf{MSE} 
        & \textbf{Params (M)} 
        & \textbf{MACs (G)} 
        & \textbf{Memory (GB)} 
        & \textbf{Train Time (s)} 
        & \textbf{Infer Time (ms)} \\
        \cmidrule(lr){1-1}
        \cmidrule(lr){2-2}
        \cmidrule(lr){3-3}
        \cmidrule(lr){4-4}
        \cmidrule(lr){5-5}
        \cmidrule(lr){6-6}
        \cmidrule(lr){7-7}
        Crossformer 
        & \underline{0.171} 
        & 36.936 
        & 47.543 
        & 22.932 
        & \underline{0.184} 
        & \underline{75.474} \\
        \cmidrule(lr){1-7}
        PatchTST    
        & \underline{0.171} 
        & \underline{5.496}  
        & \underline{23.134}  
        & \underline{12.848} 
        & 0.269 
        & 326.133 \\
        \cmidrule(lr){1-7}
        \textbf{TStruct} 
        & \textbf{0.151} 
        & \textbf{3.324} 
        & \textbf{11.779} 
        & \textbf{7.143} 
        & \textbf{0.059} 
        & \textbf{56.059} \\
        \midrule
        \textbf{Improvement ($\downarrow$)}
        & \textbf{1.13$\times$}
        & \textbf{1.65$\times$}
        & \textbf{1.96$\times$}
        & \textbf{1.80$\times$}
        & \textbf{4.56$\times$}
        & \textbf{1.35$\times$} \\
        \bottomrule[1.1pt]
    \end{tabular}
    }
\end{table}

\clearpage
\clearpage

\section{Additional Interpretability Visualizations}
\label{appendix:Interpretability}
\begin{figure}[H]
    \centering
    \includegraphics[width=1\linewidth]{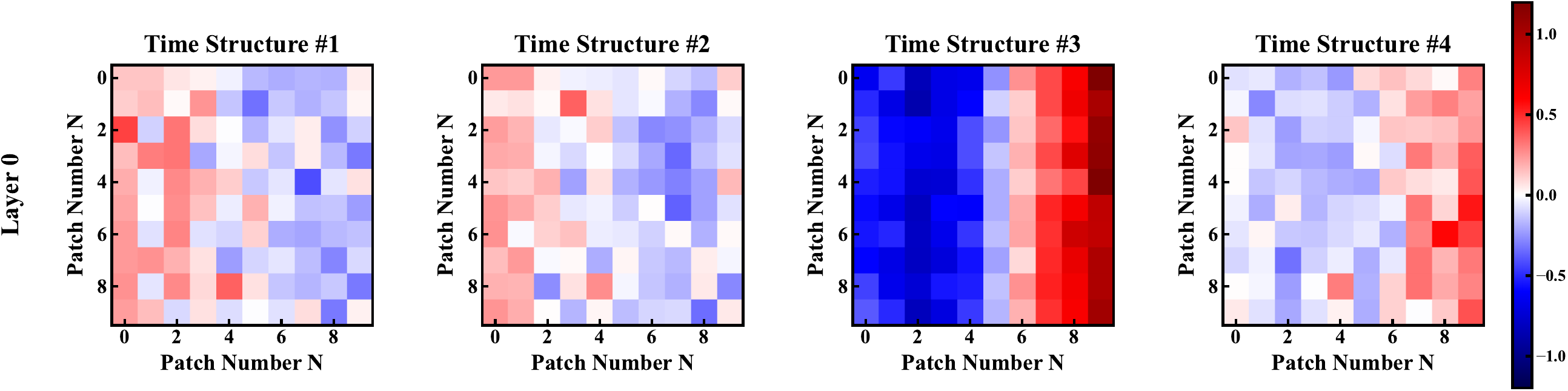}
    \caption{\textbf{Learned Temporal Structures on ETTh1.}}
    \label{fig:structure_etth1}
\end{figure}

\begin{figure}[H]
    \centering
    \includegraphics[width=1\linewidth]{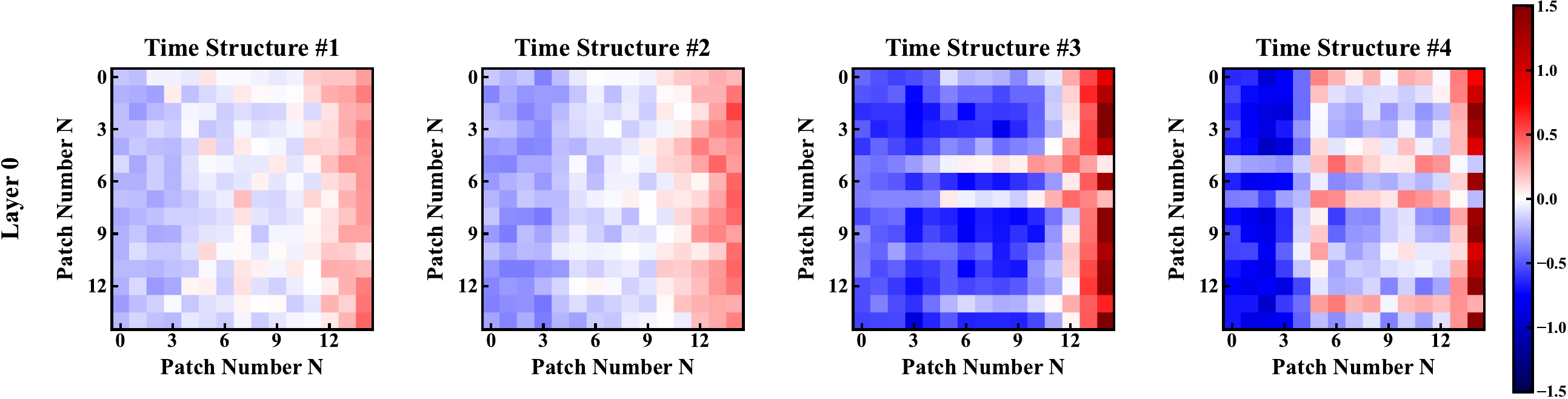}
    \caption{\textbf{Learned Temporal Structures on ETTh2.}}
    \label{fig:structure_etth2}
\end{figure}

\begin{figure}[H]
    \centering
    \includegraphics[width=1\linewidth]{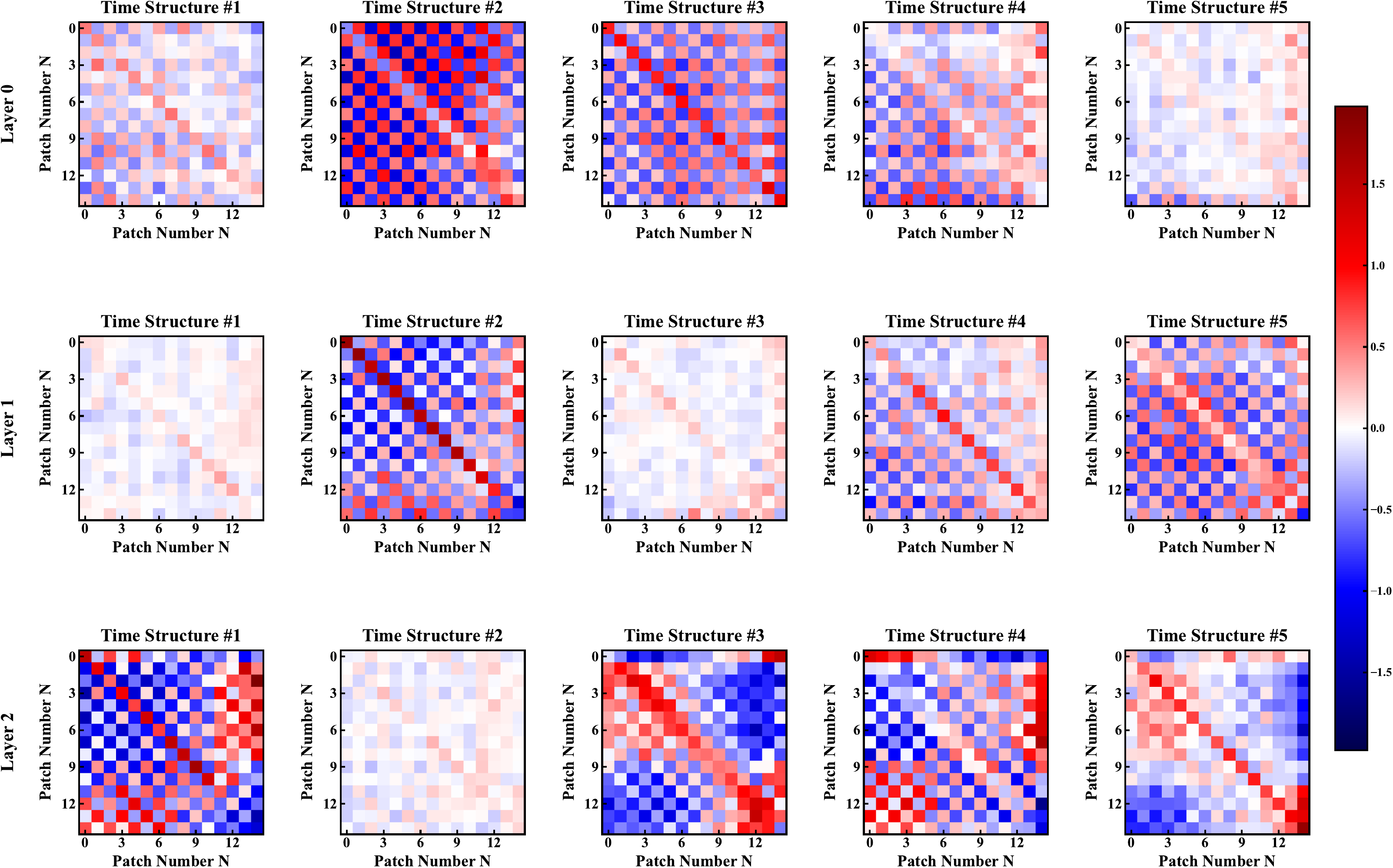}
    \caption{\textbf{Learned Temporal Structures on ETTm1.}}
    \label{fig:structure_ettm1}
\end{figure}

\begin{figure}[H]
    \centering
    \includegraphics[width=1\linewidth]{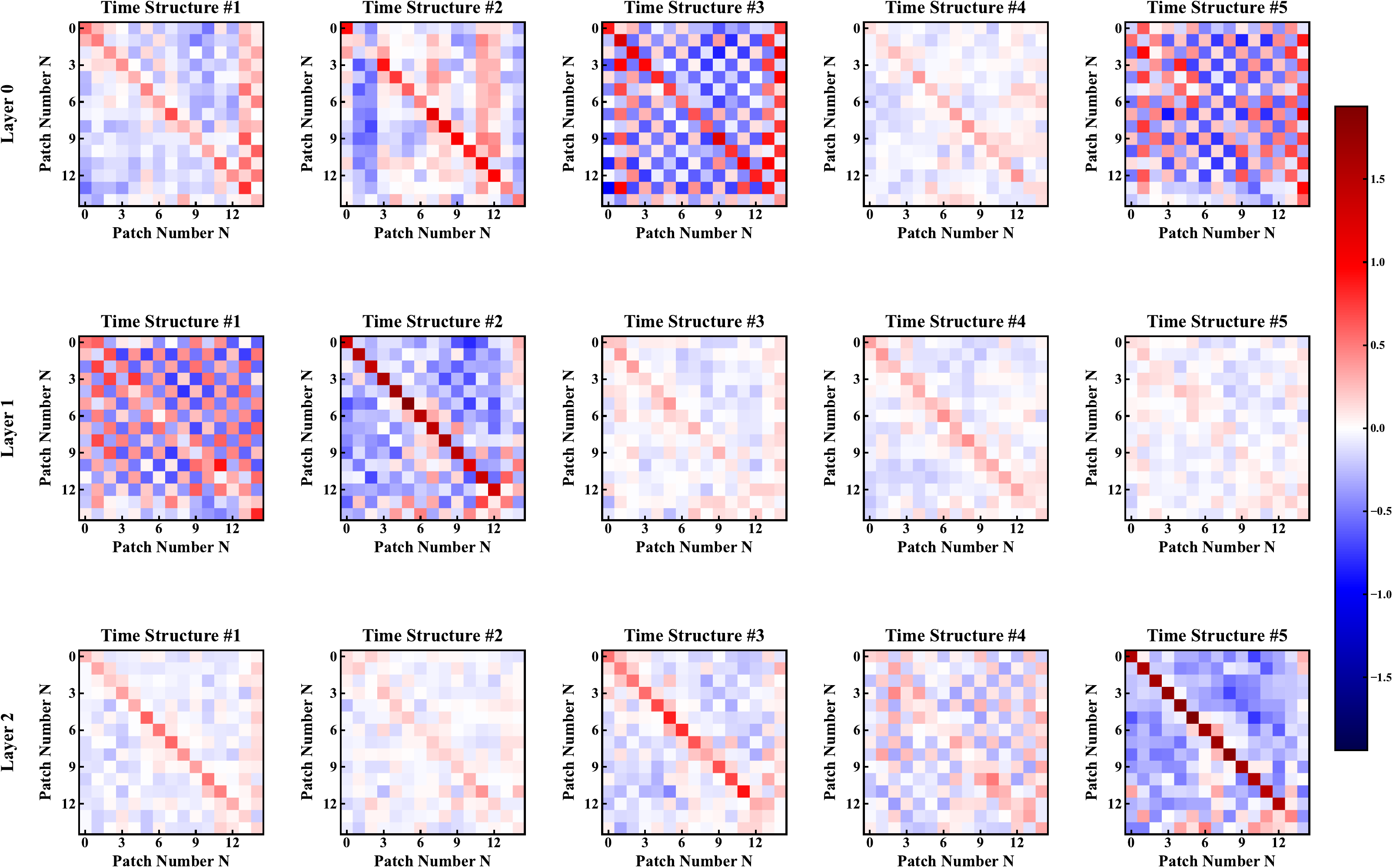}
    \caption{\textbf{Learned Temporal Structures on ETTm2.}}
    \label{fig:structure_ettm2}
\end{figure}

\begin{figure}[H]
    \centering
    \includegraphics[width=0.93\linewidth]{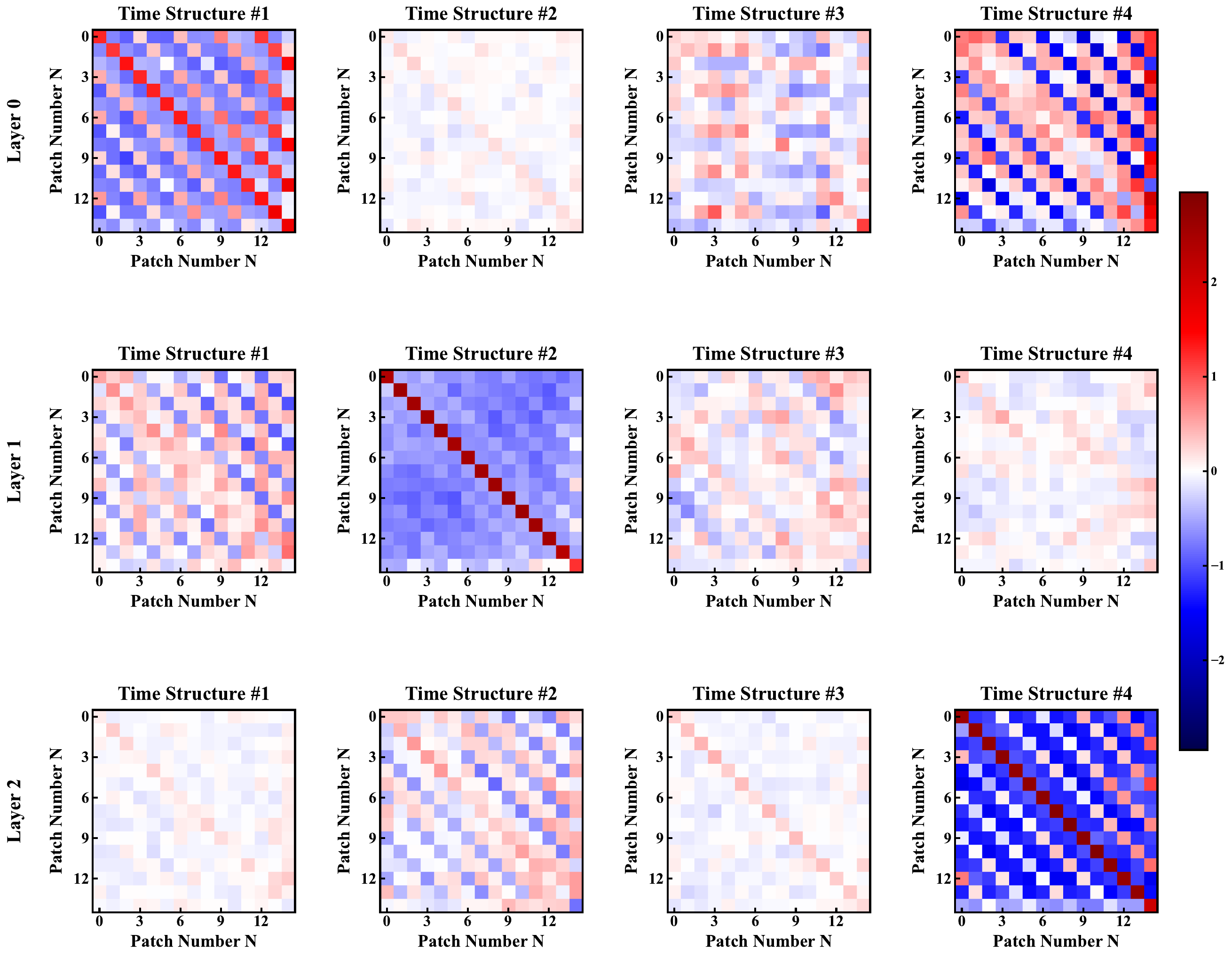}
    \caption{\textbf{Learned Temporal Structures on Weather.}}
    \label{fig:structure_weather}
\end{figure}

\begin{figure}[H]
    \centering
    \includegraphics[width=0.93\linewidth]{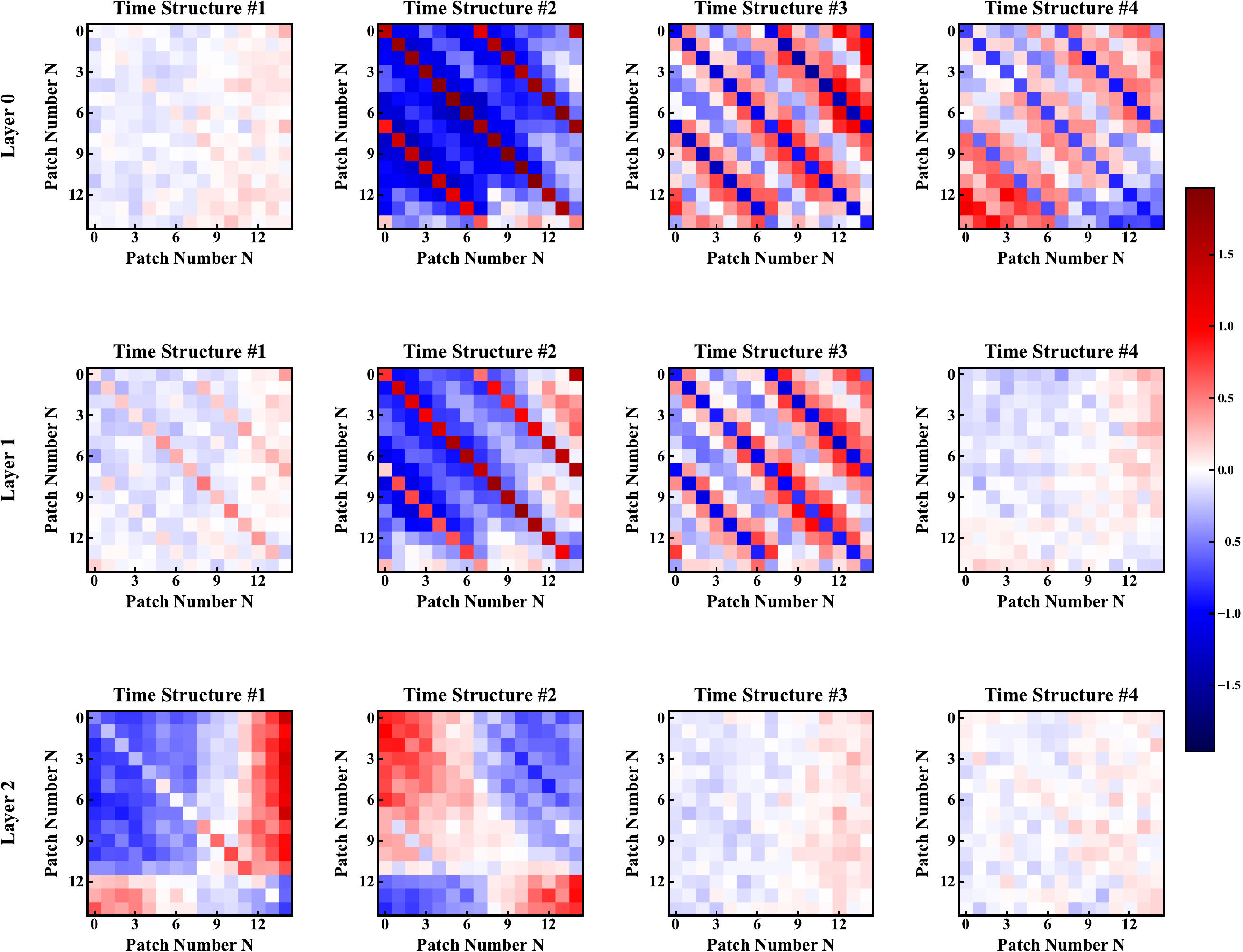}
    \caption{\textbf{Learned Temporal Structures on Electricity.}}
    \label{fig:structure_electricity}
\end{figure}

\begin{figure}[H]
    \centering
    \includegraphics[width=1\linewidth]{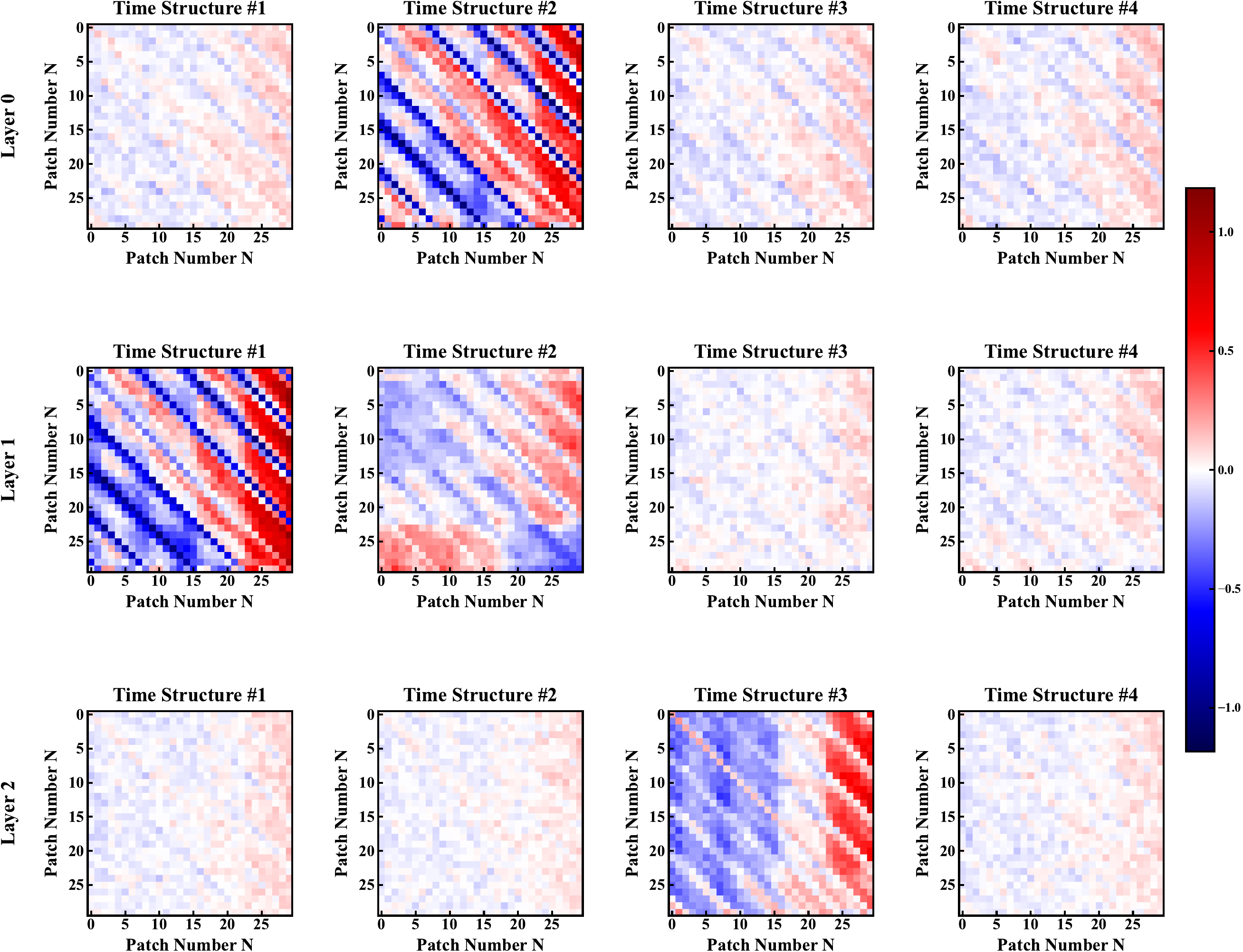}
    \caption{\textbf{Learned Temporal Structures on Traffic.}}
    \label{fig:structure_traffic}
\end{figure}

We provide additional visualizations of the learned temporal structures across all benchmark datasets. Figures~\ref{fig:structure_etth1} to~\ref{fig:structure_traffic} directly visualize the learned shared temporal structure matrices before softmax normalization, as the concentration of weights after softmax can make finer structural differences difficult to distinguish visually. Red and blue indicate higher and lower relative aggregation preferences, respectively, rather than positive and negative dependencies.
The visualizations show that the learned structures are not random mixing patterns. Instead, they exhibit structured and reusable temporal dependency patterns across datasets, including diagonal-dominant local continuity, repeated stripe-like interactions, smooth long-range transitions, and more abstract patch-level dependency patterns.

Several observations can be made. 
First, many structures contain clear diagonal or near-diagonal patterns, indicating that neighboring patches often preserve strong local temporal continuity. 
Second, repeated stripe-like or checkerboard patterns appear in several datasets, suggesting recurrent interactions between relative patch positions. 
Third, some structures exhibit smooth low-to-high transitions across rows or columns, reflecting broader temporal dependency tendencies beyond local neighborhoods. 
These patterns show that TStruct does not rely on a single predefined temporal prior, such as periodicity alone. 

The learned structures also vary across datasets. 
The ETT datasets often contain a mixture of local continuity, alternating recurrent patterns, and smooth transition structures, reflecting heterogeneous temporal dynamics. 
Weather, Electricity, and Traffic exhibit more regular banded or repeated structures, indicating that TStruct can capture dataset-specific temporal regularities through shared dependency matrices. 
Overall, these visualizations provide qualitative evidence that the temporal structures learned by TStruct are interpretable, structured, and adaptive to different datasets.
\raggedbottom
\setlength{\intextsep}{10pt}
\vspace{2em}

\vspace{2em}

\vspace{2em}

\clearpage
\section{Detailed Experimental Results}
\label{appendix:detailed_results}
This section provides the complete experimental results corresponding to the summarized tables in the main paper. 
While the main text reports average results over all prediction horizons for compactness, here we present the detailed results under each prediction length $H \in \{96,192,336,720\}$. 
These results provide a more fine-grained comparison across horizons and further support the consistency of TStruct under different forecasting settings.

\subsection{Detailed Forecasting Results}
\label{appendix:detailed_forecasting_results}
Table~\ref{tab:main_results} reports the full forecasting results of TStruct and all baseline methods on seven benchmarks. 
The results include both MSE and MAE under each prediction horizon, providing the detailed counterpart to the averaged results reported in the main paper.

\definecolor{yearblue}{rgb}{0.0, 0.0, 0.5}
\definecolor{bestredbg}{HTML}{FFCCCC}
\definecolor{secondbluebg}{HTML}{CCE5FF}
\definecolor{methodheaderbg}{HTML}{D6DCE4}
\definecolor{metricheaderbg}{HTML}{EFEFEF}

\providecommand{\redbold}[1]{}
\renewcommand{\redbold}[1]{\cellcolor{bestredbg}\textcolor{red}{\textbf{#1}}}
\providecommand{\blueunder}[1]{}
\renewcommand{\blueunder}[1]{\cellcolor{secondbluebg}\textcolor{blue}{\uline{#1}}}

\begin{table*}[htbp]
  \centering
  \caption{Full multivariate time series forecasting results. The best results are marked in \textcolor{red}{\textbf{bold red}}, and the second-best results are marked in \textcolor{blue}{\uline{blue underline}}.}
  \label{tab:main_results}
  \resizebox{\linewidth}{!}{
    \setlength{\tabcolsep}{3pt}
    \renewcommand{\arraystretch}{1.1}
    \begin{tabular}{c|c|cc|cc|cc|cc|cc|cc|cc|cc|cc}
    \toprule[1.1pt]
    \rowcolor{methodheaderbg}
    \multicolumn{2}{>{\columncolor{methodheaderbg}}c|}{\raisebox{0.5\normalbaselineskip}{\textbf{Methods}}} &
    \multicolumn{2}{>{\columncolor{methodheaderbg}}c|}{\shortstack[c]{\textbf{TStruct}\\ (ours)}} &
    \multicolumn{2}{>{\columncolor{methodheaderbg}}c|}{\shortstack[c]{\textbf{DUET}\\ \citeyearpar{duet}}} &
    \multicolumn{2}{>{\columncolor{methodheaderbg}}c|}{\shortstack[c]{\textbf{PhaseFormer}\\ \citeyearpar{phaseformer}}} &
    \multicolumn{2}{>{\columncolor{methodheaderbg}}c|}{\shortstack[c]{\textbf{GTR}\\ \citeyearpar{gtr}}} &
    \multicolumn{2}{>{\columncolor{methodheaderbg}}c|}{\shortstack[c]{\textbf{TimeEmb}\\ \citeyearpar{timeemb}}} &
    \multicolumn{2}{>{\columncolor{methodheaderbg}}c|}{\shortstack[c]{\textbf{TQNet}\\ \citeyearpar{tqnet}}} &
    \multicolumn{2}{>{\columncolor{methodheaderbg}}c|}{\shortstack[c]{\textbf{CycleNet}\\ \citeyearpar{cyclenet}}} &
    \multicolumn{2}{>{\columncolor{methodheaderbg}}c|}{\shortstack[c]{\textbf{PatchTST}\\ \citeyearpar{patchtst}}} &
    \multicolumn{2}{>{\columncolor{methodheaderbg}}c}{\shortstack[c]{\textbf{DLinear}\\ \citeyearpar{dlinear}}} \\
    \cmidrule[0.9pt](r){1-2}\cmidrule[0.9pt](lr){3-4}\cmidrule[0.9pt](lr){5-6}\cmidrule[0.9pt](lr){7-8}\cmidrule[0.9pt](lr){9-10}\cmidrule[0.9pt](lr){11-12}\cmidrule[0.9pt](lr){13-14}\cmidrule[0.9pt](lr){15-16}\cmidrule[0.9pt](lr){17-18}\cmidrule[0.9pt](l){19-20}
    \rowcolor{metricheaderbg}
    \textbf{Datasets} & \textbf{Metric} & \textbf{MSE} & \textbf{MAE} & \textbf{MSE} & \textbf{MAE} & \textbf{MSE} & \textbf{MAE} & \textbf{MSE} & \textbf{MAE} & \textbf{MSE} & \textbf{MAE} & \textbf{MSE} & \textbf{MAE} & \textbf{MSE} & \textbf{MAE} & \textbf{MSE} & \textbf{MAE} & \textbf{MSE} & \textbf{MAE} \\
    \midrule[1.1pt]
    \multirow{5}{*}{ETTh1}
      & 96  & \redbold{0.345} & \redbold{0.380} & \blueunder{0.352} & \blueunder{0.384} & 0.359 & 0.382 & 0.367 & 0.391 & 0.372 & 0.400 & 0.371 & 0.393 & 0.374 & 0.396 & 0.375 & 0.399 & 0.379 & 0.403 \\
      & 192 & \redbold{0.383} & \blueunder{0.405} & 0.398 & 0.409 & \blueunder{0.397} & \redbold{0.404} & 0.409 & 0.420 & 0.413 & 0.427 & 0.410 & 0.418 & 0.406 & 0.415 & 0.414 & 0.421 & 0.408 & 0.419 \\
      & 336 & \redbold{0.407} & \redbold{0.421} & \blueunder{0.414} & 0.426 & 0.425 & \blueunder{0.424} & 0.432 & 0.434 & 0.438 & 0.443 & 0.432 & 0.433 & 0.431 & 0.430 & 0.431 & 0.436 & 0.440 & 0.440 \\
      & 720 & \redbold{0.427} & \blueunder{0.454} & \blueunder{0.429} & 0.455 & 0.431 & \redbold{0.450} & 0.451 & 0.463 & 0.449 & 0.462 & 0.466 & 0.468 & 0.450 & 0.464 & 0.449 & 0.466 & 0.471 & 0.493 \\
      \cmidrule(lr){2-20}
      & \emph{Avg.} & \redbold{0.390} & \redbold{0.415} & \blueunder{0.398} & 0.419 & 0.403 & \redbold{0.415} & 0.415 & 0.419 & 0.418 & 0.433 & 0.420 & 0.428 & 0.415 & 0.426 & 0.417 & 0.431 & 0.425 & 0.439 \\
    \midrule
    \multirow{5}{*}{ETTh2}
      & 96  & \redbold{0.267} & \redbold{0.330} & \blueunder{0.270} & \blueunder{0.336} & 0.275 & 0.338 & 0.274 & 0.341 & 0.290 & 0.348 & 0.290 & 0.349 & 0.271 & 0.337 & 0.274 & 0.337 & 0.300 & 0.364 \\
      & 192 & \redbold{0.322} & \redbold{0.371} & \blueunder{0.332} & \blueunder{0.374} & 0.341 & 0.376 & 0.340 & 0.385 & 0.347 & 0.387 & 0.367 & 0.397 & 0.332 & 0.380 & 0.348 & 0.384 & 0.387 & 0.423 \\
      & 336 & \blueunder{0.357} & \blueunder{0.400} & \redbold{0.353} & \redbold{0.397} & 0.369 & 0.405 & 0.363 & 0.409 & 0.376 & 0.411 & 0.363 & 0.406 & 0.362 & 0.408 & 0.377 & 0.416 & 0.490 & 0.487 \\
      & 720 & \redbold{0.381} & \blueunder{0.426} & \blueunder{0.382} & \redbold{0.425} & 0.402 & 0.436 & 0.405 & 0.439 & 0.399 & 0.439 & 0.412 & 0.443 & 0.415 & 0.449 & 0.406 & 0.441 & 0.704 & 0.597 \\
      \cmidrule(lr){2-20}
      & \emph{Avg.} & \redbold{0.331} & \redbold{0.381} & \blueunder{0.334} & \blueunder{0.383} & 0.347 & 0.389 & 0.346 & 0.394 & 0.353 & 0.396 & 0.358 & 0.399 & 0.345 & 0.394 & 0.351 & 0.395 & 0.470 & 0.468 \\
    \midrule
    \multirow{5}{*}{ETTm1}
      & 96  & \redbold{0.275} & \redbold{0.325} & \blueunder{0.279} & \blueunder{0.333} & 0.293 & 0.344 & 0.283 & 0.337 & 0.293 & 0.346 & 0.294 & 0.345 & 0.299 & 0.348 & 0.289 & 0.343 & 0.300 & 0.345 \\
      & 192 & \redbold{0.320} & \redbold{0.354} & \redbold{0.320} & \blueunder{0.358} & 0.323 & 0.361 & 0.323 & 0.365 & 0.326 & 0.366 & 0.336 & 0.372 & 0.334 & 0.367 & 0.329 & 0.368 & 0.336 & 0.366 \\
      & 336 & \blueunder{0.355} & \blueunder{0.378} & \redbold{0.348} & \redbold{0.377} & 0.358 & 0.381 & 0.355 & 0.384 & 0.356 & 0.382 & 0.368 & 0.392 & 0.364 & 0.387 & 0.362 & 0.390 & 0.367 & 0.386 \\
      & 720 & \redbold{0.400} & \blueunder{0.411} & \blueunder{0.405} & \redbold{0.408} & 0.412 & 0.410 & 0.420 & 0.422 & \blueunder{0.405} & 0.410 & 0.412 & 0.443 & 0.410 & 0.411 & 0.416 & 0.423 & 0.419 & 0.416 \\
      \cmidrule(lr){2-20}
      & \emph{Avg.} & \redbold{0.337} & \redbold{0.367} & \blueunder{0.338} & \blueunder{0.369} & 0.347 & 0.374 & 0.345 & 0.377 & 0.345 & 0.376 & 0.353 & 0.388 & 0.352 & 0.378 & 0.349 & 0.381 & 0.356 & 0.378 \\
    \midrule
    \multirow{5}{*}{ETTm2}
      & 96  & \redbold{0.153} & \redbold{0.241} & \blueunder{0.161} & \blueunder{0.248} & 0.163 & 0.256 & 0.161 & 0.252 & 0.163 & 0.251 & 0.166 & 0.255 & 0.159 & 0.247 & 0.165 & 0.255 & 0.164 & 0.255 \\
      & 192 & \redbold{0.209} & \redbold{0.282} & \blueunder{0.214} & \blueunder{0.286} & 0.219 & 0.293 & 0.221 & 0.292 & 0.219 & 0.290 & 0.222 & 0.294 & 0.214 & 0.286 & 0.221 & 0.293 & 0.224 & 0.304 \\
      & 336 & \redbold{0.264} & \blueunder{0.321} & \blueunder{0.267} & \blueunder{0.321} & 0.269 & 0.326 & 0.278 & 0.326 & \blueunder{0.265} & \redbold{0.320} & 0.277 & 0.337 & 0.268 & 0.326 & 0.276 & 0.327 & 0.277 & 0.337 \\
      & 720 & \blueunder{0.344} & \redbold{0.372} & 0.348 & 0.374 & 0.351 & 0.379 & 0.359 & 0.388 & \redbold{0.343} & \redbold{0.372} & 0.366 & 0.386 & 0.353 & 0.384 & 0.362 & 0.381 & 0.371 & 0.401 \\
      \cmidrule(lr){2-20}
      & \emph{Avg.} & \redbold{0.243} & \redbold{0.304} & \blueunder{0.248} & \blueunder{0.307} & 0.251 & 0.314 & 0.255 & 0.315 & \blueunder{0.248} & 0.308 & 0.258 & 0.318 & 0.249 & 0.311 & 0.256 & 0.314 & 0.259 & 0.324 \\
    \midrule
    \multirow{5}{*}{Weather}
      & 96  & \redbold{0.137} & \redbold{0.181} & 0.146 & \blueunder{0.191} & 0.148 & 0.195 & 0.147 & 0.199 & \blueunder{0.143} & 0.193 & 0.155 & 0.201 & 0.164 & 0.220 & 0.150 & 0.200 & 0.170 & 0.230 \\
      & 192 & \redbold{0.181} & \redbold{0.226} & \blueunder{0.188} & \blueunder{0.231} & 0.193 & 0.237 & 0.192 & 0.240 & \blueunder{0.188} & 0.237 & 0.200 & 0.242 & 0.209 & 0.258 & 0.191 & 0.239 & 0.216 & 0.273 \\
      & 336 & \redbold{0.230} & \redbold{0.267} & \blueunder{0.234} & \blueunder{0.268} & 0.242 & 0.278 & 0.243 & 0.281 & 0.236 & 0.274 & 0.248 & 0.284 & 0.255 & 0.292 & 0.242 & 0.279 & 0.258 & 0.307 \\
      & 720 & \redbold{0.304} & \blueunder{0.324} & \blueunder{0.305} & \redbold{0.319} & 0.309 & 0.332 & 0.324 & 0.337 & 0.306 & 0.325 & 0.318 & 0.335 & 0.320 & 0.338 & 0.312 & 0.330 & 0.323 & 0.362 \\
      \cmidrule(lr){2-20}
      & \emph{Avg.} & \redbold{0.213} & \redbold{0.249} & \blueunder{0.218} & \blueunder{0.252} & 0.223 & 0.261 & 0.227 & 0.264 & \blueunder{0.218} & 0.257 & 0.230 & 0.266 & 0.237 & 0.277 & 0.224 & 0.262 & 0.242 & 0.293 \\
    \midrule
    \multirow{5}{*}{Electricity}
      & 96  & \redbold{0.123} & \redbold{0.216} & 0.128 & \blueunder{0.219} & 0.129 & 0.221 & \blueunder{0.127} & 0.222 & 0.129 & 0.225 & 0.129 & 0.224 & 0.128 & 0.223 & 0.130 & 0.222 & 0.140 & 0.237 \\
      & 192 & \redbold{0.141} & \redbold{0.232} & 0.145 & \blueunder{0.235} & 0.148 & 0.238 & 0.147 & 0.242 & 0.145 & 0.241 & 0.151 & 0.245 & \blueunder{0.143} & 0.237 & 0.148 & 0.240 & 0.154 & 0.251 \\
      & 336 & \redbold{0.155} & \redbold{0.250} & 0.163 & 0.255 & 0.165 & 0.257 & 0.165 & 0.260 & 0.161 & 0.257 & 0.166 & 0.261 & \blueunder{0.159} & \blueunder{0.254} & 0.167 & 0.261 & 0.169 & 0.268 \\
      & 720 & \redbold{0.185} & \redbold{0.279} & \blueunder{0.193} & \blueunder{0.281} & 0.201 & 0.285 & 0.199 & 0.292 & 0.197 & 0.289 & 0.201 & 0.294 & 0.197 & 0.287 & 0.202 & 0.291 & 0.204 & 0.301 \\
      \cmidrule(lr){2-20}
      & \emph{Avg.} & \redbold{0.151} & \redbold{0.244} & \blueunder{0.157} & \blueunder{0.248} & 0.161 & 0.250 & 0.160 & 0.254 & 0.158 & 0.253 & 0.162 & 0.256 & 0.157 & 0.250 & 0.162 & 0.254 & 0.167 & 0.264 \\
    \midrule
    \multirow{5}{*}{Traffic}
      & 96  & \redbold{0.352} & \redbold{0.235} & \blueunder{0.360} & 0.238 & 0.361 & \blueunder{0.238} & 0.367 & 0.257 & 0.374 & 0.268 & 0.362 & 0.260 & 0.381 & 0.266 & 0.367 & 0.251 & 0.395 & 0.275 \\
      & 192 & \redbold{0.368} & \blueunder{0.244} & 0.383 & 0.249 & \blueunder{0.373} & \redbold{0.243} & 0.382 & 0.266 & 0.387 & 0.268 & 0.384 & 0.269 & 0.394 & 0.273 & 0.385 & 0.259 & 0.407 & 0.280 \\
      & 336 & \redbold{0.383} & \blueunder{0.253} & 0.395 & 0.259 & \blueunder{0.385} & \redbold{0.248} & 0.399 & 0.275 & 0.401 & 0.275 & 0.423 & 0.297 & 0.406 & 0.279 & 0.398 & 0.265 & 0.417 & 0.286 \\
      & 720 & \redbold{0.422} & \blueunder{0.277} & 0.435 & 0.278 & \blueunder{0.428} & \redbold{0.270} & 0.441 & 0.298 & 0.434 & 0.275 & 0.447 & 0.303 & 0.441 & 0.300 & 0.434 & 0.287 & 0.454 & 0.308 \\
      \cmidrule(lr){2-20}
      & \emph{Avg.} & \redbold{0.381} & \blueunder{0.252} & 0.393 & 0.256 & \blueunder{0.387} & \redbold{0.250} & 0.397 & 0.274 & 0.399 & 0.272 & 0.404 & 0.282 & 0.406 & 0.280 & 0.396 & 0.266 & 0.418 & 0.287 \\
    \midrule
    \multicolumn{2}{c|}{\textbf{1st Count}} & \redbold{32} & \redbold{23} & \blueunder{3} & \textbf{5} & \textbf{0} & \blueunder{7} & \textbf{0} & \textbf{0} & \textbf{1} & \textbf{2} & \textbf{0} & \textbf{0} & \textbf{0} & \textbf{0} & \textbf{0} & \textbf{0} & \textbf{0} & \textbf{0} \\
    \bottomrule[1.1pt]
    \end{tabular}
  }
\end{table*}

\subsection{Detailed Ablation Results}
\label{appendix:detailed_ablation_results}

Table~\ref{tab:ablation_results} reports the full ablation results under different prediction horizons. 
These results complement the averaged ablation table in the main paper and show that the benefits of the proposed Temporal Structure Mixer and PCAHead remain consistent across forecasting horizons.

\definecolor{headergray}{HTML}{D6DCE4}
\definecolor{metricgray}{HTML}{EFEFEF}

\begin{table*}[!t]
\centering
\caption{
Complete ablation results of TStruct under different prediction horizons.
TSM$\rightarrow$MHA replaces the Temporal Structure Mixer with multi-head attention,
TSM$\rightarrow$Linear replaces it with a shared linear temporal mixer,
and PCAHead$\rightarrow$Linear replaces the proposed prediction head with a linear head.
}
\label{tab:ablation_results}
\vspace{0.03in}

\footnotesize
\setlength{\tabcolsep}{3.8pt}
\renewcommand{\arraystretch}{1.10}

\resizebox{0.99\textwidth}{!}{
\begin{tabular}{>{\centering\arraybackslash}m{1.15cm}|>{\centering\arraybackslash}m{0.9cm}|cc|cc|cc|cc|cc|cc|cc}
\toprule[1.1pt]
\rowcolor{headergray}
\multicolumn{2}{c|}{\textbf{Datasets}}
& \multicolumn{2}{c|}{\textbf{ETTh1}}
& \multicolumn{2}{c|}{\textbf{ETTh2}}
& \multicolumn{2}{c|}{\textbf{ETTm1}}
& \multicolumn{2}{c|}{\textbf{ETTm2}}
& \multicolumn{2}{c|}{\textbf{Weather}}
& \multicolumn{2}{c|}{\textbf{Electricity}}
& \multicolumn{2}{c}{\textbf{Traffic}} \\
\cmidrule[0.9pt](r){1-2}
\cmidrule[0.9pt](lr){3-4}
\cmidrule[0.9pt](lr){5-6}
\cmidrule[0.9pt](lr){7-8}
\cmidrule[0.9pt](lr){9-10}
\cmidrule[0.9pt](lr){11-12}
\cmidrule[0.9pt](lr){13-14}
\cmidrule[0.9pt](l){15-16}
\rowcolor{metricgray}
\textbf{Method}
& \textbf{Metric}
& \textbf{MSE} & \textbf{MAE}
& \textbf{MSE} & \textbf{MAE}
& \textbf{MSE} & \textbf{MAE}
& \textbf{MSE} & \textbf{MAE}
& \textbf{MSE} & \textbf{MAE}
& \textbf{MSE} & \textbf{MAE}
& \textbf{MSE} & \textbf{MAE} \\
\midrule[1.0pt]

\multirow{5}{*}{\shortstack[c]{TSM\\$\rightarrow$ MHA}}
& 96  & 0.367 & 0.399 & 0.285 & 0.350 & 0.298 & 0.340 & 0.191 & 0.272 & 0.143 & 0.189 & 0.131 & 0.225 & 0.377 & 0.255 \\
& 192 & 0.421 & 0.440 & 0.360 & 0.395 & 0.344 & 0.369 & 0.222 & 0.293 & 0.188 & 0.232 & 0.148 & 0.240 & 0.390 & 0.262 \\
& 336 & 0.434 & 0.447 & 0.360 & 0.402 & 0.424 & 0.426 & 0.266 & 0.322 & 0.260 & 0.295 & 0.167 & 0.258 & 0.405 & 0.272 \\
& 720 & 0.478 & 0.491 & 0.393 & 0.433 & 0.437 & 0.429 & 0.356 & 0.383 & 0.326 & 0.344 & 0.203 & 0.291 & 0.439 & 0.289 \\
\cmidrule[0.8pt](lr){2-16}
& \emph{Avg.} & 0.425 & 0.444 & 0.350 & 0.395 & 0.376 & 0.391 & 0.259 & 0.318 & 0.229 & 0.265 & 0.162 & 0.254 & 0.403 & 0.270 \\
\midrule

\multirow{5}{*}{\shortstack[c]{TSM\\$\rightarrow$ Linear}}
& 96  & 0.364 & 0.393 & 0.271 & 0.327 & 0.278 & 0.331 & 0.158 & 0.243 & 0.140 & 0.185 & 0.125 & 0.218 & 0.352 & 0.235 \\
& 192 & 0.395 & 0.413 & 0.338 & 0.373 & 0.334 & 0.362 & 0.218 & 0.285 & 0.187 & 0.232 & 0.143 & 0.234 & 0.368 & 0.245 \\
& 336 & 0.418 & 0.428 & 0.365 & 0.404 & 0.359 & 0.382 & 0.266 & 0.322 & 0.236 & 0.273 & 0.157 & 0.250 & 0.383 & 0.254 \\
& 720 & 0.433 & 0.461 & 0.391 & 0.431 & 0.418 & 0.413 & 0.351 & 0.381 & 0.309 & 0.327 & 0.187 & 0.280 & 0.423 & 0.277 \\
\cmidrule[0.8pt](lr){2-16}
& \emph{Avg.} & 0.403 & 0.424 & 0.341 & 0.384 & 0.347 & 0.372 & 0.248 & 0.308 & 0.218 & 0.254 & 0.153 & 0.246 & 0.382 & 0.253 \\
\midrule

\multirow{5}{*}{\shortstack[c]{PCAHead\\$\rightarrow$ Linear}}
& 96  & 0.348 & 0.382 & 0.268 & 0.331 & 0.275 & 0.325 & 0.160 & 0.248 & 0.137 & 0.179 & 0.124 & 0.217 & 0.353 & 0.236 \\
& 192 & 0.385 & 0.405 & 0.330 & 0.374 & 0.323 & 0.354 & 0.212 & 0.283 & 0.184 & 0.227 & 0.142 & 0.232 & 0.369 & 0.245 \\
& 336 & 0.411 & 0.423 & 0.364 & 0.405 & 0.357 & 0.379 & 0.267 & 0.322 & 0.230 & 0.268 & 0.156 & 0.250 & 0.383 & 0.254 \\
& 720 & 0.431 & 0.459 & 0.393 & 0.432 & 0.408 & 0.412 & 0.354 & 0.376 & 0.310 & 0.328 & 0.186 & 0.279 & 0.423 & 0.277 \\
\cmidrule[0.8pt](lr){2-16}
& \emph{Avg.} & 0.394 & 0.417 & 0.339 & 0.386 & 0.341 & 0.368 & 0.248 & 0.307 & 0.215 & 0.251 & 0.152 & 0.245 & 0.382 & 0.253 \\
\midrule

\multirow{5}{*}{\textbf{TStruct}}
& 96  & \textbf{0.345} & \textbf{0.380} & \textbf{0.267} & \textbf{0.330} & \textbf{0.275} & \textbf{0.325} & \textbf{0.155} & \textbf{0.241} & \textbf{0.137} & \textbf{0.181} & \textbf{0.123} & \textbf{0.216} & \textbf{0.352} & \textbf{0.235} \\
& 192 & \textbf{0.383} & \textbf{0.405} & \textbf{0.322} & \textbf{0.371} & \textbf{0.320} & \textbf{0.354} & \textbf{0.209} & \textbf{0.282} & \textbf{0.181} & \textbf{0.226} & \textbf{0.141} & \textbf{0.232} & \textbf{0.368} & \textbf{0.244} \\
& 336 & \textbf{0.407} & \textbf{0.421} & \textbf{0.357} & \textbf{0.400} & \textbf{0.355} & \textbf{0.378} & \textbf{0.265} & \textbf{0.320} & \textbf{0.230} & \textbf{0.267} & \textbf{0.155} & \textbf{0.249} & \textbf{0.383} & \textbf{0.253} \\
& 720 & \textbf{0.427} & \textbf{0.454} & \textbf{0.381} & \textbf{0.426} & \textbf{0.400} & \textbf{0.411} & \textbf{0.344} & \textbf{0.372} & \textbf{0.304} & \textbf{0.324} & \textbf{0.185} & \textbf{0.279} & \textbf{0.422} & \textbf{0.277} \\
\cmidrule[0.8pt](lr){2-16}
& \emph{Avg.} & \textbf{0.390} & \textbf{0.416} & \textbf{0.331} & \textbf{0.381} & \textbf{0.337} & \textbf{0.367} & \textbf{0.243} & \textbf{0.303} & \textbf{0.213} & \textbf{0.249} & \textbf{0.151} & \textbf{0.244} & \textbf{0.381} & \textbf{0.252} \\
\bottomrule[1.1pt]
\end{tabular}

}
\end{table*}

\clearpage
\section{Limitations and Future Directions}
\label{appendix:limitations}
TStruct models temporal dependencies through a compact set of shared temporal structure matrices, with controlled initialization and softmax normalized composition and aggregation. This parameterization provides an effective way to learn reusable temporal structures for LTSF. Their adaptation across patch resolutions and transfer across datasets and domains at scale remain directions for further exploration.

\textbf{Patch resolution dependency.}
The learned temporal structures are tied to a specific patch resolution. Since each temporal dependency matrix is parameterized over $N$ patch positions, changing the patch size or look-back length can alter the matrix dimensions or the temporal meaning of its coordinates. Although TStruct remains effective under different look-back lengths, directly sharing temporal structures across different patch resolutions remains non-trivial. A promising direction is to develop resolution-adaptive temporal structures that can be transferred, interpolated, or composed across different patch scales.

\textbf{Temporal structure foundation models.}
Even after fine-tuning, time series foundation models can struggle to outperform small models trained on individual datasets. One possible explanation is the difficulty of learning input representations that generalize across heterogeneous time series, motivating the exploration of temporal dependency structures as an alternative focus of pretraining. Our analysis provides evidence of reusable temporal relationships across temporal positions and variables, while their transfer and generalization across datasets and domains at scale remain untested. A potential direction is to develop a time series foundation model built around a pretrained dictionary of shared temporal dependency structures, with downstream forecasting tasks selecting and composing these structures to form temporal aggregation operators.

\end{document}